%% file: main.tex
\documentclass[10pt,twocolumn,letterpaper]{article}

\usepackage[pagenumbers]{cvpr} 

\input{preamble}

\definecolor{cvprblue}{rgb}{0.21,0.49,0.74}

\usepackage[pagebackref,breaklinks,colorlinks,citecolor=cvprblue]{hyperref}

\usepackage{makecell}
\usepackage{colortbl}
\usepackage{color}
\definecolor{linkcolor}{RGB}{255,0,0}
\definecolor{urlcolor}{RGB}{255,105,180}
\definecolor{citecolor}{RGB}{66,168,235}
\usepackage{rotating} 
\usepackage{pifont}
\usepackage{multirow}
\usepackage{multicol}
\usepackage{lineno}
\usepackage{marvosym}
\hypersetup{colorlinks=true,linkcolor=linkcolor,urlcolor=urlcolor,citecolor=blue}
\definecolor{tab_others}{RGB}{235, 235, 235}
\definecolor{tab_ours}{RGB}{225, 235, 246}
\newcommand{\envelope}{\ding{41}} 

\def\paperID{646} 
\def\confName{CVPR}
\def\confYear{2025}

\title{Dinomaly: The \textit{Less Is More} Philosophy in \\ Multi-Class Unsupervised Anomaly Detection}

\author{%
Jia Guo\textsuperscript{1}~~~~Shuai Lu\textsuperscript{2}~~~~Weihang Zhang\textsuperscript{2 \dag}~~~~Fang Chen\textsuperscript{3}~~~~Huiqi Li\textsuperscript{2 \dag}~~~~Hongen Liao\textsuperscript{1,3 \envelope } \\
\\
\textsuperscript{1}School of Biomedical Engineering, Tsinghua University, Beijing, China\\
\textsuperscript{2}School of Information and Electronics, Beijing Institute of Technology, Beijing, China\\
\textsuperscript{3}School of Biomedical Engineering, Shanghai Jiao Tong University, Shanghai, China\\
{\small \texttt{guojia.jeremy@gmail.com}~~~\texttt{lushuaie@163.com}~~~\texttt{zhangweihang@bit.edu.cn}
}
\\
{\small ~~~\texttt{chen-fang@sjtu.edu.cn} ~~~\texttt{huiqili@bit.edu.cn}~~~\texttt{liao@tsinghua.edu.cn}~~~
}
\\
}

\begin{document}
\maketitle

\renewcommand{\thefootnote}{}  
\noindent\footnotetext{Accepted by CVPR 2025. \envelope~Corresponding author. \dag~Advisors when the project initiated.}
\renewcommand{\thefootnote}{\arabic{footnote}}  

\input{sec/0_abstract}    
\input{sec/1_intro}
\input{sec/2_relatedwork}

\input{sec/3_method}

\input{sec/4_experiments}

\section*{Acknowledgments}
The authors acknowledge supports from National Natural Science Foundation
of China (U22A2051, 82027807), National
Key Research and Development Program of China
(2022YFC2405200), Tsinghua-Foshan Innovation Special Fund (2021THFS0104), and Institute for
Intelligent Healthcare, Tsinghua University (2022ZLB001).

{
    \small
    \bibliographystyle{ieeenat_fullname}
    \bibliography{main}
}

\input{sec/X_suppl}

\end{document}

%% file: preamble.tex
\usepackage[dvipsnames]{xcolor}


%% file: sec/0_abstract.tex
\begin{abstract}
  Recent studies highlighted a practical setting of unsupervised anomaly detection (UAD) that builds a unified model for multi-class images. Despite various advancements addressing this challenging task, the detection performance under the multi-class setting still lags far behind state-of-the-art class-separated models. Our research aims to bridge this substantial performance gap. In this paper, we present Dinomaly, a minimalist reconstruction-based anomaly detection framework that harnesses pure Transformer architectures without relying on complex designs, additional modules, or specialized tricks. Given this powerful framework consisting of only Attentions and MLPs, we found four simple components that are essential to multi-class anomaly detection: (1) Scalable foundation Transformers that extracts universal and discriminative features, (2) Noisy Bottleneck where pre-existing Dropouts do all the noise injection tricks, (3) Linear Attention that naturally cannot focus, and (4) Loose Reconstruction that does not force layer-to-layer and point-by-point reconstruction. Extensive experiments are conducted across popular anomaly detection benchmarks including MVTec-AD, VisA, Real-IAD, etc. Our proposed Dinomaly achieves impressive image-level AUROC of \textbf{99.6}\%, \textbf{98.7}\%, and \textbf{89.3}\% on the three datasets respectively, which is not only superior to state-of-the-art multi-class UAD methods, but also achieves the most advanced class-separated UAD records. 
  Code is available at: \url{https://github.com/guojiajeremy/Dinomaly}
\end{abstract}

%% file: sec/1_intro.tex
\section{Introduction}
\label{sec:intro}

Unsupervised anomaly detection (UAD) aims to detect abnormal patterns from normal images and further localize the anomalous regions. Because of the diversity of potential anomalies and their scarcity, this task is proposed to model the accessible training sets containing only normal samples as an unsupervised paradigm. UAD has a wide range of applications, e.g., industrial defect detection \cite{bergmann2019mvtec}, medical disease screening \cite{guo2023encoder}, and video surveillance \cite{mabrouk2018abnormal}, addressing the difficulty of collecting and labeling all possible anomalies in these scenarios.

\begin{figure*}[!t]
\centerline{\includegraphics[width=0.96\textwidth]{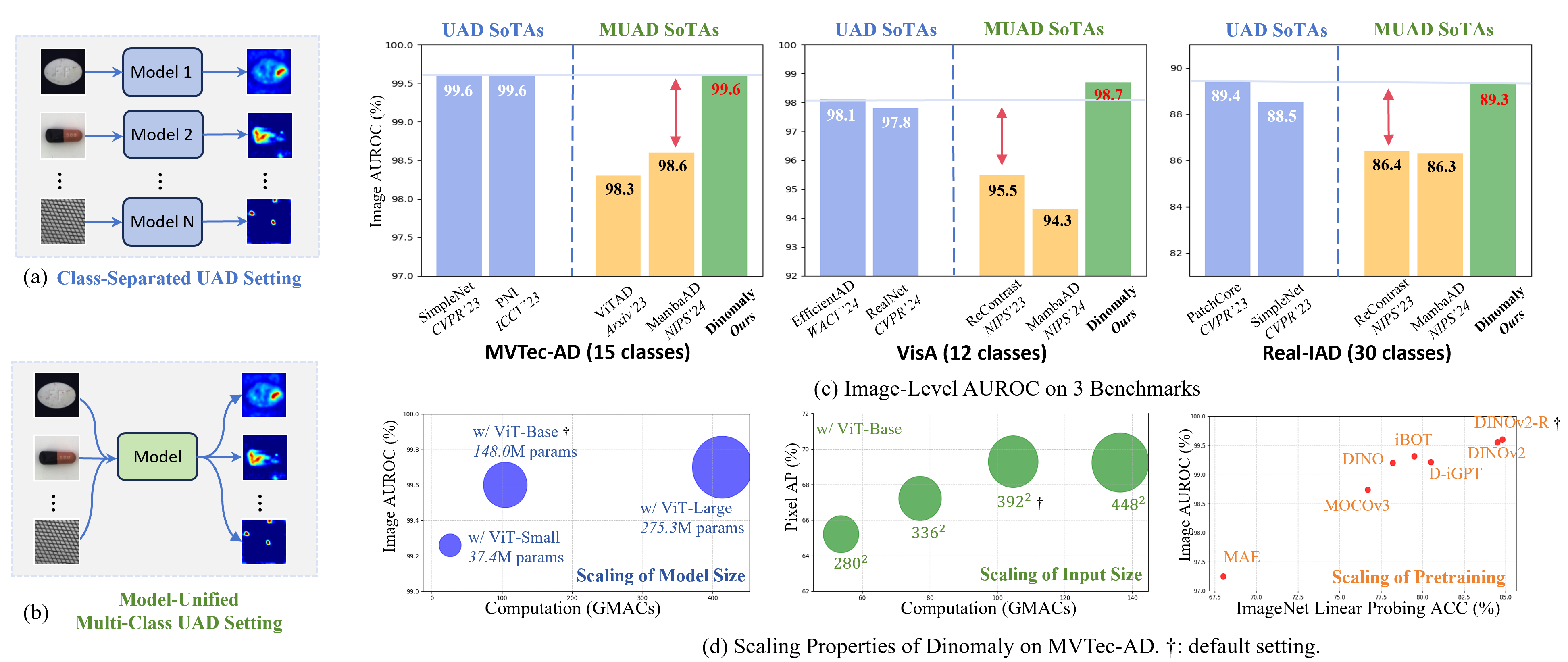}}
\caption{Setting, benchmarking, and scaling of Dinomaly. (a) Task setting of class-separated UAD. (b) Task setting of MUAD. (c) Comparison with previous SoTA methods on MVTec-AD \cite{bergmann2019mvtec}, VisA \cite{zou2022spot}, and Real-IAD \cite{wang2024real}. (d) Scaling properties of Dinomaly.}
\label{fig1}
\end{figure*}

Conventional works on UAD build a separate model for each object category, as shown in Figure~\ref{fig1}(a). However, this one-class-one-model setting entails substantial storage overhead for saving models \cite{you2022unified}, especially when the application scenario necessitates a large number of object classes. For UAD methods, a compact boundary of normal patterns is vital to distinguish anomalies. Once the intra-normal patterns become exceedingly complicated due to various classes, the corresponding distribution becomes challenging to measure, consequently harming the detection performance. Recently, UniAD \cite{you2022unified} and successive studies have been proposed to train a unified model for multi-class anomaly detection (MUAD), as shown in Figure~\ref{fig1}(b). Under this setting, the "identity mapping" that directly copies the input as the output regardless of normal or anomaly harms the performance of conventional methods \cite{you2022unified}. This phenomenon is caused by the diversity of multi-class normal patterns that drive the network to generalize on unseen patterns. 

Within two years, a number of methods have been proposed to address MUAD, such as neighbor-masked attention \cite{you2022unified}, synthetic anomalies \cite{zhao2023omnial}, vector quantization \cite{lu2023hierarchical}, diffusion model \cite{yin2023lafite,he2023diad}, and state space model (Mamba) \cite{he2024mambaad}. However, there is still a non-negligible performance gap between the state-of-the-art (SoTA) MUAD methods and class-separated UAD methods, restricting the practicability of implementing unified models, as shown in Figure~\ref{fig1}(c). In addition, previous methods employ modules and architectures delicately designed, which may not be straightforward, and consequently suffer from limited universality and ease-of-use \cite{he2024diffusion,lu2023hierarchical}. 

In this work, we aim to catch up with the performance of class-separated anomaly detection models using a multi-class unified model. We introduce Dinomaly, a minimalist reconstruction-based UAD framework built exclusively by pure Transformer blocks \cite{vaswani2017attention}, specifically Self-Attentions and Multi-Layer Perceptrons (MLPs). \textit{To begin with}, we empirically investigate the scaling law of self-supervised pre-trained Vision Transformers (ViT) \cite{dosovitskiy2020image} when serving as the feature encoders for extracting reconstruction objectives. Subsequently, we introduce three straightforward yet crucial elements to address the critical identity mapping phenomenon in MUAD contexts, without increasing complexity or computational burden. \textit{First}, as an alternative to meticulously designed pseudo anomaly and feature noise, we propose to activate the built-in Dropout in an MLP to prevent the network from restoring both normal and anomalous patterns. \textit{Second}, we propose to leverage the "side effect" of Linear Attention (a computation-efficient counterpart of Softmax Attention) that impedes focus on local regions, thus preventing the forwarding of identical information. \textit{Third}, previous methods adopt layer-to-layer and region-by-region reconstruction schemes, distilling a decoder that can well mimic the encoder's behavior even for anomalous regions. Therefore, we propose to loosen the reconstruction constraints by grouping multiple layers as a whole and discarding well-reconstructed regions during optimization.

To validate the effectiveness of the proposed Dinomaly under MUAD setting, we conduct extensive experiments on a number of widely used benchmarks, including MVTec AD \cite{bergmann2019mvtec} (15 classes), VisA \cite{zou2022spot} (12 classes), and Real-IAD (30 classes). As shown in Figure~\ref{fig1}, our base-size Dinomaly achieves unprecedented image-level AUROC of \textbf{99.6}\%, \textbf{98.7}\%, and \textbf{89.3}\% on MVTec AD, VisA, and Real-IAD, surpassing previous SoTA methods by a large margin. In addition, scalability is a key feature of Dinomaly. Further scaling up the model size maximizes performance to the fullest level of \textbf{99.8}\%, \textbf{98.9}\%, and \textbf{90.1}\%, respectively; while scaling down parameters and input size can offer efficient solutions in computation-constrained scenarios.

%% file: sec/2_relatedwork.tex
\section{Related Work}
\label{sec:related}

\begin{figure*}[!t]
\centerline{\includegraphics[width=\textwidth]{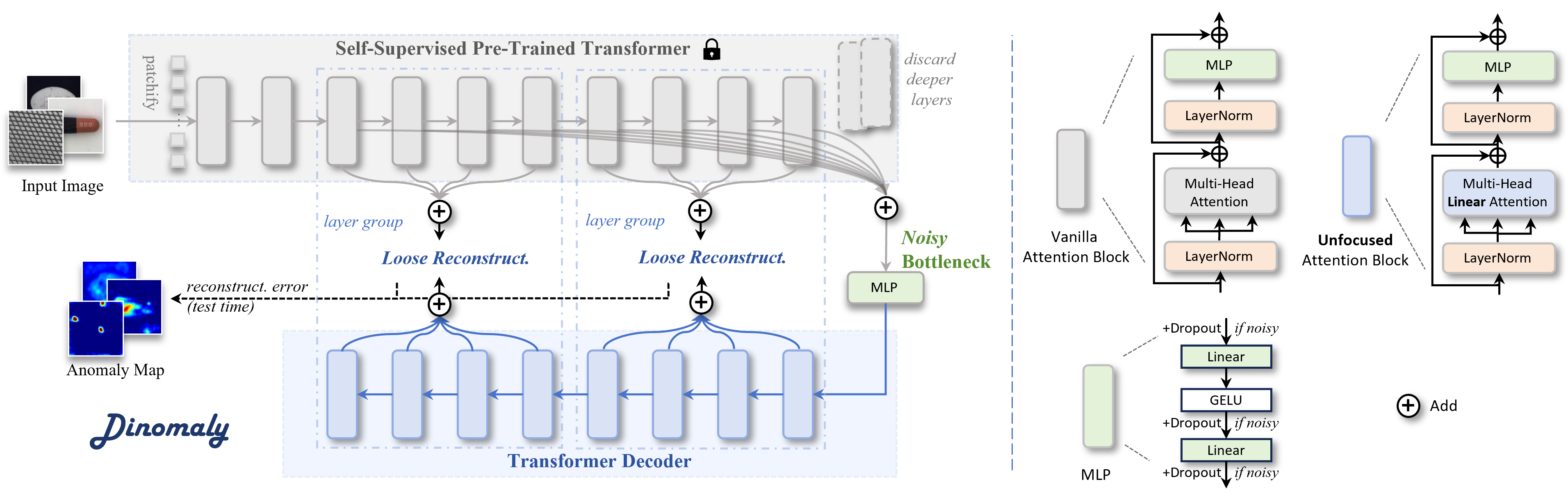}}
\caption{The framework of Dinomaly, built by simple and pure Transformer building blocks.}
\label{fig2}
\end{figure*}

\textbf{Multi-Class UAD}. UniAD \cite{you2022unified} first introduced multi-class anomaly detection, aiming to detect anomalies for different classes using a unified model. In this setting, conventional UAD methods often face the challenge of "identical shortcuts", where both anomaly-free and anomaly samples can be effectively recovered during inference \cite{you2022unified}. It is believed that this phenomenon is caused by the diversity of multi-class normal patterns that drive the network to generalize on unseen patterns. This contradicts the fundamental assumption of epistemic methods. Many current researches focus on addressing this challenge \cite{you2022unified,lu2023hierarchical,guo2023recontrast,liu2023mixed,yin2023lafite}. UniAD \cite{you2022unified} employs a neighbor-masked attention module and a feature-jitter strategy to mitigate these shortcuts. HVQ-Trans \cite{lu2023hierarchical} proposes a vector quantization (VQ)  Transformer model that induces large feature discrepancies for anomalies. LafitE \cite{yin2023lafite} utilizes a latent diffusion model and introduces a feature editing strategy to alleviate this issue. DiAD \cite{he2023diad} also employs diffusion models to address multi-class UAD settings. OmniAL \cite{zhao2023omnial} focuses on anomaly localization in the unified setting, preventing identical reconstruction by using synthesized pseudo anomalies. ReContrast \cite{guo2023recontrast} attempted to alleviate the identity mapping by cross-reconstruction between two encoders. ViTAD \cite{chen2021exploring} abstracts a unified feature-reconstruction UAD framework and employ Transformer building blocks. MambaAD \cite{he2024mambaad} explores the recently proposed State Space Model (SSM), Mamba, in the context of multi-class UAD. More related works of UAD are presented in Appendix~\ref{sec:relatedwork}.

%% file: sec/3_method.tex
\section{Method}

\subsection{Dinomaly Framework}
\textit{“What I cannot create, I do not understand”}

------Richer Feynman

The ability to recognize anomalies from what we know is an innate human capability, serving as a vital pathway for us to explore the world. Similarly, we construct a reconstruction-based framework that relies on the epistemic characteristic of artificial neural networks. Dinomaly consists of an encoder, a bottleneck, and a reconstruction decoder, as shown in Figure~\ref{fig2}. Without loss of generality, a pretrained ViT network \cite{dosovitskiy2020image} with 12 Transformer layers is used as the encoder, extracting informative feature maps with different semantic scales. The bottleneck is a simple MLP (a.k.a. feed-forward network, FFN) that collects the feature representations of the encoder's 8 middle-level layers. The decoder is similar to the encoder, consisting of 8 Transformer layers. During training, the decoder learns to reconstruct the middle-level features of the encoder by maximizing the cosine similarity between feature maps. During inference, the decoder is expected to reconstruct normal regions of feature maps but fails for anomalous regions as it has never seen such samples. 

\textbf{Foundation Transformers.}
Foundation models, especially ViTs \cite{dosovitskiy2020image,liu2021swin} pre-trained on large-scale datasets, serve as a basis and starting point for specific computer vision tasks. Such networks employ self-supervised learning schemes such as contrastive learning (MoCov3 \cite{chen2021empirical}, DINO \cite{caron2021emerging}), masked image modeling (MAE \cite{he2022masked}, SimMIM \cite{xie2022simmim}, BEiT \cite{peng2002beitv2}), and their combination (iBOT \cite{zhou2021ibot}, DINOv2 \cite{oquab2023dinov2}), producing universal features suitable for image-level visual tasks and pixel-level visual tasks. 
 
Because of the lack of supervision in UAD, most advanced methods adopt pre-trained networks to extract discriminative features. Recent works \cite{reiss2022anomaly,zhang2023exploring,lee2023selformaly} have preliminarily discovered the advantage of robust and universal features of self-supervised models over domain-specific ImageNet features in anomaly detection tasks.  In this paper, we pioneer the investigation of scaling behaviors in UAD models through a systematic analysis of foundational ViTs, as briefed in Figure~\ref{fig1}(d). Our comprehensive evaluation encompasses pre-training strategies (Figure~\ref{fig:backbone}), model sizes (Table~\ref{tab:arch}), and input resolutions (Table~\ref{tab:inputsize}), which are detailed in section~\ref{sec:ab}. Considering the balance of detection performance and computational efficiency, we adopt ViT-Base/14 pretrained by DINOv2-Register \cite{darcet2023vision} as the encoder of Dinomaly by default.

\begin{figure*}[!t]
\centerline{\includegraphics[width=0.85\textwidth]{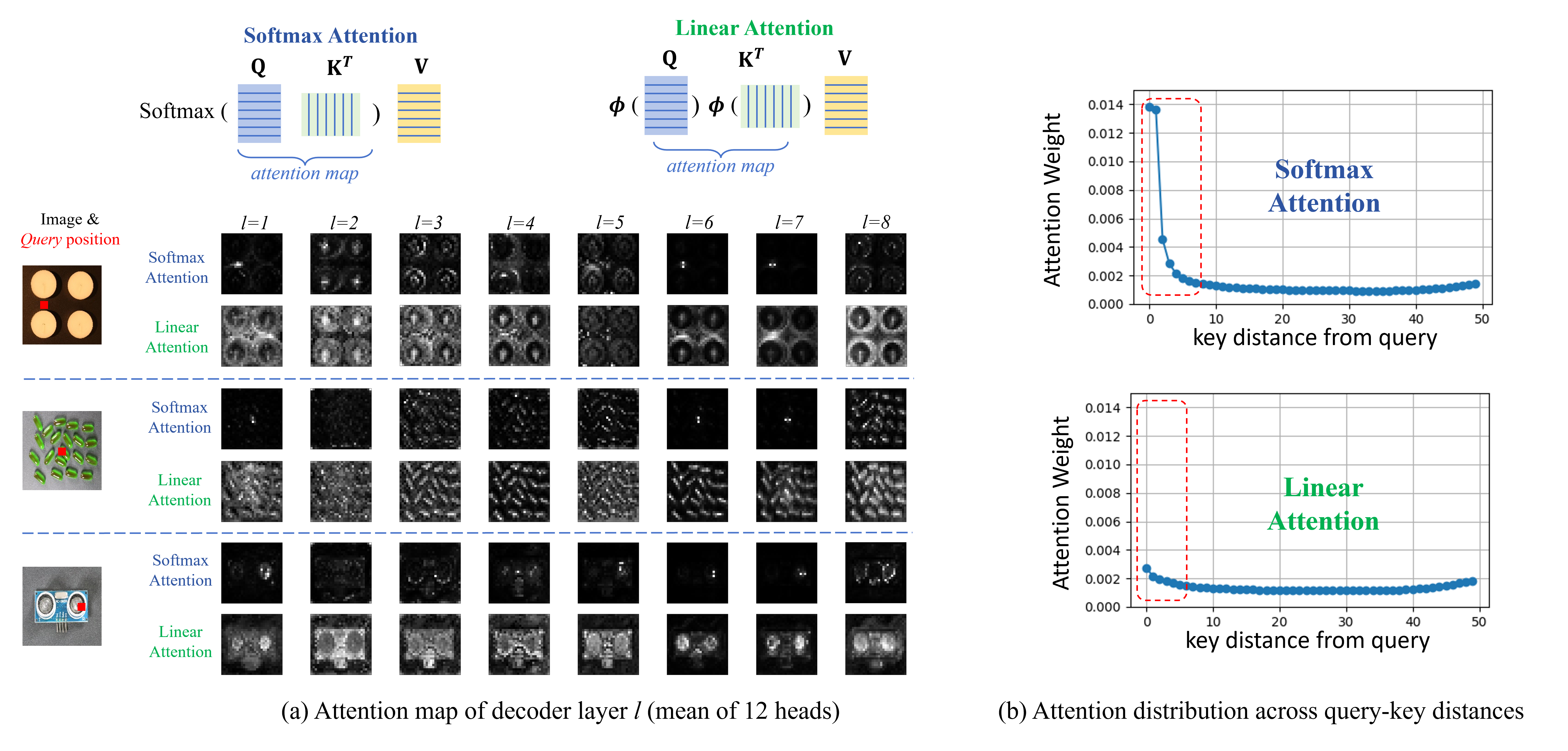}}
\caption{Softmax Attention \textit{vs.} Linear Attention. (a) Visualization of attention maps. (b) Attention distribution.}
\label{fig3}
\end{figure*}

\subsection{Noisy Bottleneck.}
\textit{“Dropout is all you need.”}

Prior studies \cite{you2022unified,zhao2023omnial,guo2023recontrast} attribute the performance degradation of UAD methods trained on diverse multi-class samples to the "identity mapping" phenomenon; in this work, we reframe this issue as an "over-generalization" problem. Generalization ability is a merit of neural networks, allowing them to perform equally well on unseen test sets. However, generalization is not so wanted in the context of unsupervised anomaly detection that leverages the epistemic nature of neural networks. With the increasing diversity of images and their patterns due to multi-class UAD settings, the decoder can generalize its reconstruction ability to unseen anomalous samples, resulting in the failure of anomaly detection using reconstruction error.

A direct solution for identity mapping is to shift "reconstruction" to "restoration". Specifically, instead of directly reconstructing the normal images or features given normal inputs, previous works propose to add perturbations as pseudo anomalies on input images \cite{zavrtanik2021draem,zhang2023destseg} or on encoder features \cite{you2022unified,yin2023lafite}; meanwhile, still let the decoder restore anomaly-free images or features, formulating a denoising-like framework. However, such methods employ heuristic and hand-crafted anomaly generation strategies that may not be universal across domains, datasets, and methods. In this work, we turn to leveraging the simple and elegant Dropout techniques. Since its introduction by Hinton et al. \cite{hinton2012dropout} in 2014 as a remedy for overfitting, Dropout has become a cornerstone element in neural architectures, including Transformers. In Dinomaly, we employ Dropout to randomly discard neural activations in an MLP bottleneck. Instead of alleviating overfitting, the role of Dropout in Dinomaly can be explained as pseudo feature anomaly that perturb normal representations, analogous to denoising auto-encoders \cite{vincent2008extracting,vincent2010stacked}. Without introducing any specific modules, this simple component inherently forces the decoder to restore normal features regardless of whether the test image contains anomalies, in turn, mitigating identical mapping.

\subsection{Unfocused Linear Attention.}
\textit{“One man's poison is another man's meat”}

\textbf{Softmax Attention} is the key mechanism of Transformers, allowing the model to attend to different parts of its input token sequence. Formally, given an input sequence $\mathbf{X} \in \mathbb{R}^{N \times d}$ with length $N$, Attention first transforms it into three matrices: the query matrix $\mathbf{Q} \in \mathbb{R}^{N \times d}$, the key matrix $\mathbf{K} \in \mathbb{R}^{N \times d}$, and the value matrix $\mathbf{V} \in \mathbb{R}^{N \times d}$:
\begin{equation}
\mathbf{Q} = \mathbf{X} \mathbf{W}^Q ~,
\mathbf{K} = \mathbf{X} \mathbf{W}^K ~,
\mathbf{V} = \mathbf{X} \mathbf{W}^V ~,
\end{equation}
where $\mathbf{W}^Q, \mathbf{W}^K, \mathbf{W}^V \in \mathbb{R}^{d \times d}$ are learnable parameters. By computing the attention map by the query-key similarity, the output of Softmax Attention is given as:  \footnote{The full form of Attention is $\text{Softmax}(\frac{\mathbf{Q}\mathbf{K}^T}{\sqrt{d}}) \mathbf{V}$. The constant denominator is omitted for narrative simplicity. The multi-head mechanism that concatenates multiple Attentions is also omitted.}
\begin{equation}
\text{Attention}(\mathbf{Q}, \mathbf{K}, \mathbf{V})= \text{Softmax}(\mathbf{Q}\mathbf{K}^T) \mathbf{V} ~.
\end{equation}

Back to MUAD, previous methods \cite{you2022unified,lu2023hierarchical} suggest adopting Attentions instead of Convolutions because Convolutions can easily learn identical mappings. Nevertheless, both operations are in danger of forming identity mapping by over-concentrating on corresponding input locations for producing the outputs:
\begin{align*}
\text{Conv Kernel} &=
\begin{bmatrix}
0 & 0 & 0\\
0 & 1 & 0\\
0 & 0 & 0
\end{bmatrix} ~, &
\text{Attn Map} &=
\begin{bmatrix}
1 & 0 & 0\\
0 & 1 & 0\\
0 & 0 & 1\\
\end{bmatrix} ~.
\end{align*}

\begin{figure*}[!t]
\centerline{\includegraphics[width=0.98\textwidth]{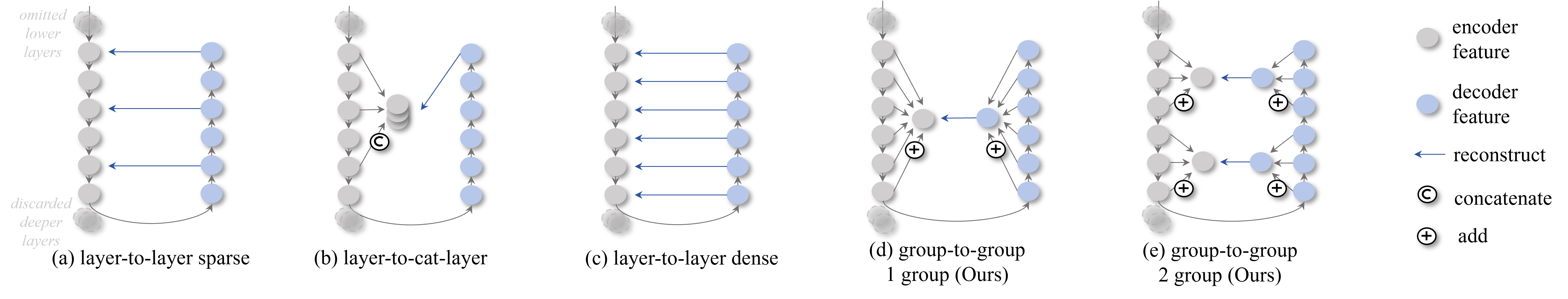}}
\caption{Schemes of reconstruction constraint. (a) Layer-to-layer (sparse). (b) Layer-to-cat-layer. (c) Layer-to-layer (dense).  (d) Loose group-to-group, 1-group (Ours). (e) Loose group-to-group, 2-group (Ours). }
\label{fig4}
\end{figure*}

Is there any simple solution that prevents Attentions from attending to identical information? In Dinomaly,  we turn to leverage the "unfocusing ability" of a type of Softmax-free Attention, i.e., \textbf{Linear Attention}. Linear Attention was proposed as a promising alternative to reduce the computation complexity of vanilla Softmax Attention concerning the number of tokens \cite{katharopoulos2020transformers}. By substituting Softmax operation with a simple activation function $\phi( \cdot )$ (usually $\phi(x) = \text{elu}(x)+1$), we can change the computation order from $(\mathbf{Q} \mathbf{K}^T)\mathbf{V}$ to $\mathbf{Q} (\mathbf{K}^T \mathbf{V})$. Formally, Linear Attention (LA) is given as:
\begin{equation}
\text{LA}(\mathbf{Q}, \mathbf{K}, \mathbf{V})= (\phi(\mathbf{Q}) \phi(\mathbf{K}^T)) \mathbf{V} = \phi(\mathbf{Q})(\phi(\mathbf{K}^T) \mathbf{V}) ~,
\end{equation}
where the computation complexity is reduced to $\mathcal{O}(Nd^2)$ from $\mathcal{O}(N^2d)$. The trade-off between complexity and expressiveness is a dilemma. Previous studies \cite{han2023flatten,shen2021efficient} attribute Linear Attention's performance degradation on supervised tasks to its incompetence in focusing. Due to the absence of non-linear attention reweighting by Softmax operation, Linear Attention cannot concentrate on important regions related to the query, such as foreground and neighbors. This property, however, is exactly what the reconstruction decoder favors in our contexts.

In order to probe how Attentions propagate information, we train two variants of Dinomaly using vanilla Softmax Attention or Linear Attention as the spatial mixer in the decoder and visualize their attention maps. As shown in Figure \ref{fig3}, Softmax Attention tends to focus on the exact region of the query, while Linear Attention spreads its attention across the whole image. This implies that Linear Attention, forced by its incompetence to focus, utilizes more long-range information to restore features at each position, reducing the chance of passing identical information of unseen patterns to the next layer during reconstruction. Of course, employing Linear Attention also benefits from less computation, free of performance drop.

\begin{table*}[t]
  \centering
  \tiny
  \caption{Performance under \textbf{multi-class} UAD setting (\%). \dag: method designed for MUAD. \textcolor{gray}{Dinomaly$\uparrow$}: training schedule is scaled up to 20,000, 20,000, and 100,000 iterations (original: 10,000/10,000/50,000).}
   \resizebox{0.9\textwidth}{!}{
    \begin{tabular}{ccccccccc}
    \toprule
    \multirow{2}[2]{*}{Dateset} & \multirow{2}[2]{*}{Method} & \multicolumn{3}{c}{Image-level} & \multicolumn{4}{c}{Pixel-level} \\
\cmidrule(r){3-5} \cmidrule(l){6-9} 
& &\multicolumn{1}{c}{AUROC} & \multicolumn{1}{c}{AP} & \multicolumn{1}{c}{$F_1$-max} & \multicolumn{1}{c}{AUROC} & \multicolumn{1}{c}{AP} & \multicolumn{1}{c}{$F_1$-max} & \multicolumn{1}{c}{AUPRO} \\
\hline
    \multirow{9}[0]{*}{\textbf{MVTec-AD}~\cite{bergmann2019mvtec}} & RD4AD~\cite{deng2022anomaly} & 94.6  & 96.5  & 95.2  & 96.1  & 48.6  & 53.8  & 91.1 \\
          & SimpleNet~\cite{liu2023simplenet} & 95.3  & 98.4  & 95.8  & 96.9  & 45.9  & 49.7  & 86.5\\
          & DeSTSeg~\cite{zhang2023destseg} & 89.2  & 95.5  & 91.6  & 93.1  & 54.3  & 50.9  & 64.8\\
          & UniAD~\cite{you2022unified}\dag & 96.5  & 98.8  & 96.2  & 96.8  & 43.4  & 49.5  & 90.7\\
          & ReContrast~\cite{guo2023recontrast}\dag & 98.3 & 99.4 & 97.6 & 97.1 & \underline{60.2} & \underline{61.5} & \underline{93.2} \\
          & DiAD~\cite{he2024diffusion}\dag  & 97.2  & 99.0  & 96.5  & 96.8  & 52.6  & 55.5  & 90.7\\
          & ViTAD~\cite{zhang2023exploring}\dag & 98.3 & 99.4 & 97.3 & \underline{97.7} & 55.3 & 58.7 & 91.4 \\
          & MambaAD~\cite{he2024mambaad}\dag  & \underline{98.6} & \underline{99.6} & \underline{97.8} & \underline{97.7} & 56.3 & 59.2 & 93.1  \\
          & \textbf{Dinomaly} (Ours) &\textbf{99.6} & \textbf{99.8} & \textbf{99.0} & \textbf{98.4} & \textbf{69.3} & \textbf{69.2} & \textbf{94.8} \\
        & \textcolor{gray}{Dinomaly$\uparrow$} & \textcolor{gray}{99.7} & \textcolor{gray}{99.8} & \textcolor{gray}{99.2} & \textcolor{gray}{98.4} & \textcolor{gray}{69.3} & \textcolor{gray}{69.2} & \textcolor{gray}{94.7} \\
        
    \hline
    \multirow{9}[0]{*}{\textbf{VisA}~\cite{zou2022spot}} & RD4AD~\cite{deng2022anomaly} & 92.4  & 92.4  & 89.6 & 98.1 & 38.0  & 42.6  &  91.8 \\
          & SimpleNet~\cite{liu2023simplenet} & 87.2  & 87.0  & 81.8  & 96.8  & 34.7  & 37.8  & 81.4\\
          & DeSTSeg~\cite{zhang2023destseg} & 88.9  & 89.0  & 85.2  & 96.1  & 39.6 & 43.4  & 67.4\\
          & UniAD~\cite{you2022unified}\dag & 88.8  & 90.8  & 85.8  & 98.3  & 33.7  & 39.0  & 85.5\\
          & ReContrast~\cite{guo2023recontrast}\dag & \underline{95.5} & \underline{96.4} & \underline{92.0} & \underline{98.5} & \underline{47.9} & \underline{50.6} & \underline{91.9}  \\
          & DiAD~\cite{he2024diffusion}\dag  & 86.8  & 88.3  & 85.1  & 96.0  & 26.1  & 33.0  & 75.2\\
          & ViTAD~\cite{zhang2023exploring}\dag & 90.5 & 91.7 &86.3 & 98.2 & 36.6 & 41.1 & 85.1 \\
          & MambaAD~\cite{he2024mambaad}\dag  & 94.3 & 94.5 & 89.4  & \underline{98.5} & 39.4  & 44.0 & 91.0  \\
          & \textbf{Dinomaly} (Ours) & \textbf{98.7} & \textbf{98.9} & \textbf{96.2} & \textbf{98.7} & \textbf{53.2} & \textbf{55.7} & \textbf{94.5} \\
        & \textcolor{gray}{Dinomaly$\uparrow$} & \textcolor{gray}{98.9} & \textcolor{gray}{99.0} & \textcolor{gray}{96.4} & \textcolor{gray}{98.8} & \textcolor{gray}{53.8} & \textcolor{gray}{55.8} & \textcolor{gray}{94.5} \\
        
    \hline
    \multirow{9}[0]{*}{\textbf{Real-IAD}~\cite{wang2024real}}& RD4AD~\cite{deng2022anomaly}&82.4  & 79.0  & 73.9  & 97.3  & 25.0  & 32.7  & 89.6 \\
        & SimpleNet~\cite{liu2023simplenet}&57.2  & 53.4  & 61.5  & 75.7  & \;\;2.8   & \;\;6.5   & 39.0 \\
        & DeSTSeg~\cite{zhang2023destseg}&82.3  & 79.2  & 73.2  & 94.6  & \underline{37.9} & \underline{41.7} & 40.6 \\
        & UniAD~\cite{you2022unified}\dag & 83.0  & 80.9  & 74.3  & 97.3  & 21.1  & 29.2  & 86.7\\
        & ReContrast~\cite{guo2023recontrast}\dag & \underline{86.4}  & 84.2  & \underline{77.4}  & 97.8  & 31.6  & 38.2  & \underline{91.8} \\
        & DiAD~\cite{he2024diffusion}\dag &75.6  & 66.4  & 69.9  & 88.0  &  \;\;2.9   &  \;\;7.1   & 58.1 \\
        & ViTAD~\cite{zhang2023exploring}\dag & 82.7& 80.2 & 73.7 & 97.2 & 24.3 & 32.3 & 84.8 \\
        & MambaAD~\cite{he2024mambaad}\dag &86.3 & \underline{84.6} & 77.0 & \underline{98.5} & 33.0  & 38.7  & 90.5 \\
        & \textbf{Dinomaly} (Ours) & \textbf{89.3} & \textbf{86.8} & \textbf{80.2} & \textbf{98.8} & \textbf{42.8} & \textbf{47.1} & \textbf{93.9}  \\
        & \textcolor{gray}{Dinomaly$\uparrow$} & \textcolor{gray}{89.5} & \textcolor{gray}{86.9} & \textcolor{gray}{80.4} & \textcolor{gray}{98.9} & \textcolor{gray}{43.3} & \textcolor{gray}{47.4} & \textcolor{gray}{94.2} \\
        
    \bottomrule
    \end{tabular}%
    }
  \label{tab1}%
\end{table*}%

\subsection{Loose Reconstruction}
\textit{“The tighter you squeeze, the less you have.”}

\textbf{Loose Constraint.} Pioneers of feature-reconstruction/distillation UAD methods  \cite{salehi2021multiresolution,deng2022anomaly} are inspired by knowledge distillation \cite{hinton2015distilling}. Most reconstruction-based methods distill specific encoder layers (e.g. 3 last layers of 3 ResNet stages) by the corresponding decoder layers \cite{deng2022anomaly,salehi2021multiresolution,zhang2023exploring} (Figure \ref{fig4}(a)) or the last decoder layer \cite{you2022unified,yang2020dfr} (Figure \ref{fig4}(b)). Intuitively, with more encoder-decoder feature pairs (Figure \ref{fig4}(c)), UAD model can utilize more information in different layers to discriminate anomalies. However, according to the intuition of knowledge distillation, the student (decoder) can better mimic the behavior of the teacher (encoder) given more layer-to-layer supervision \cite{liang2024module}, which is harmful for UAD models that detect anomalies by encoder-decoder discrepancy. This phenomenon is also embodied as identity mapping. Thanks to the top-to-bottom consistency of columnar Transformer layers, we propose to loosen the layer-to-layer constraint by adding up all feature maps of interested layers as a whole group, as shown in Figure \ref{fig4}(d). This scheme can be seen as loosening the layer-to-layer correspondence and providing the decoder with more degrees of freedom, so that the decoder is allowed to act much more differently from the encoder when the input pattern is unseen. Because features of shallow layers contain low-level visual characters that are helpful for precise localization, we can further group the features into the low-semantic-level group and high-semantic-level group, as shown in Figure \ref{fig4}(e).

\textbf{Loose Loss.} Following the above analysis, we also loosen the point-by-point reconstruction loss function by discarding some points in the feature map. Here, we simply borrow the hard-mining global cosine loss \cite{guo2023recontrast} that detaches the gradients of well-restored feature points with low cosine distance during training. Let $f_{E}$ and $f_{D}$ denotes (grouped) feature maps of encoder and decoder:
\begin{equation}
\mathcal{L}_{global-hm} = \mathcal{D}_{cos}(\mathcal{F}( f_{E} ), \mathcal{F}( \hat{f_{D}} )),
\end{equation}
\begin{equation}
\small
\hat{f_{D}}(h,w) = \left\{ \begin{array}{l}
sg( f_{D}( h,w ) )_{0.1}, \text{ if } \mathcal{D}_{cos}( f_{D}, f_{E} ) < k\%_{batch}\\
f_{D}( h,w ), \text{ else} \\
\end{array} \right.
\label{eq:ghm}
\end{equation}
\begin{equation}
\small
\mathcal{D}_{cos}(a, b) = 1 - \frac{{a}^{T} \cdot b}{\left\| {a} \right\| ~\left\| {b} \right\|},
\end{equation}

where $\mathcal{D}_{cos}$ denotes cosine distance, $\mathcal{F}(\cdot)$ denotes flatten operation, $f_{D}(h,w)$ represents the feature point at $(h,w)$, and $sg(\cdot)_{0.1}$ denotes shrink the gradient to one-tenth of the original \footnote{Complete stop-gradient causes optimization instability occasionally.}. $\mathcal{D}_{cos}( f_{D}( h,w ), f_{E}( h,w ) )<k\%_{batch}$ selects $k\%$ feature points with smaller cosine distance within a batch for gradient shrinking. Total loss is the average $\mathcal{L}_{global-hm}$ of all encoder-decoder feature pairs.

%% file: sec/4_experiments.tex
\section{Experiments}
\subsection{Experimental Settings}

\textbf{Datasets.} \textbf{MVTec-AD} \cite{bergmann2019mvtec} contains 15 objects (5 texture classes and 10 object classes) with a total of 3,629 normal images as the training set and 1,725 images as the test set (467 normal, 1,258 anomalous). \textbf{VisA} \cite{zou2022spot} contains 12 objects. Training and test sets are split following the official splitting, resulting in 8,659 normal images in the training set and 2,162 images in the test set (962 normal, 1,200 anomalous). \textbf{Real-IAD} \cite{wang2024real} is a large UAD dataset recently released, containing 30 distinct objects. We follow the official splitting that includes all views, resulting in 36,465 normal images in the training set and 114,585 images in the test set (63,256 normal, 51,329 anomalous). 

\textbf{Metrics.} Following prior works \cite{he2024mambaad,zhang2023exploring}, we adopt 7 evaluation metrics. Image-level anomaly detection performance is measured by the Area Under the Receiver Operator Curve (AUROC), Average Precision (AP), and $F_{1}$ score under optimal threshold ($F_{1}$-max).  Pixel-level anomaly localization is measured by AUROC, AP, $F_{1}$-max and the Area Under the Per-Region-Overlap (AUPRO). The results of a dataset is the average of all classes.

\begin{table*}[t]
  \centering
  \scriptsize
  \caption{Performance under conventional \textbf{class-separated} UAD setting (\%). n/a: not available.}
   \resizebox{0.9\linewidth}{!}{
     \begin{tabular}{@{}cccccccccc@{}}
    \toprule
    \multirow{2}[2]{*}{Method}  & \multicolumn{3}{c}{MVTec-AD~\cite{bergmann2019mvtec}} & \multicolumn{3}{c}{VisA~\cite{zou2022spot}} & \multicolumn{3}{c}{Real-IAD~\cite{wang2024real}} \\
\cmidrule(r){2-4} \cmidrule(r){5-7} \cmidrule(l){8-10} & I-AUROC       & P-AUROC          & P-AUPRO       & I-AUROC       & P-AUROC          & P-AUPRO       & I-AUROC       & P-AUROC & P-AUPRO \\ \midrule
\textit{\textbf{Dinomaly} (MUAD)}          & \textit{99.6}& \textit{98.4}& \textit{94.8}& \textit{98.7}& \textit{98.7} & \textit{94.5}& \textit{89.3}          & \textit{98.8} & \textit{93.9}    \\ \midrule

\textbf{Dinomaly}                 & \textbf{99.7} & \textbf{99.9}& \textbf{95.0}& \textbf{98.9} & \textbf{98.9}    & \textbf{95.1}& \textbf{92.0} & \textbf{99.1} & \textbf{95.1}    \\
RD4AD~\cite{deng2022anomaly}                   & 98.5          & 97.8          & \underline{93.9} & 96.0          & 90.1          & 70.9          & 87.1          & n/a& \underline{93.8}    \\
PatchCore~\cite{roth2022towards}   & 99.1          & \underline{98.1}          & 93.5          & 94.7          & \underline{98.5}          & 91.8          & \underline{89.4}          & n/a& 91.5    \\
SimpleNet~\cite{liu2023simplenet}       & \underline{99.6}& \underline{98.1}          & 90.0          & \underline{97.1}          & 98.2          & \underline{90.7}          & 88.5          & n/a & 84.6    \\
\bottomrule
\end{tabular}
}
  \label{tab2}%
\end{table*}

\begin{table*}[!t]
  \centering
    \tiny
  \caption{Ablations of Dinomaly elements on MVTec-AD (\%). NB: Noisy Bottleneck. LA: Linear Attention. LC: Loose Constraint (2 groups). LL: Loose Loss. As MVTec-AD has reached saturation, we also present the results on VisA (Table~\ref{tab:abvisa}).}
   \resizebox{0.9\linewidth}{!}{
    \begin{tabular}{ccccccccccc}
    \toprule
    \multirow{2}[2]{*}{NB} & \multirow{2}[2]{*}{LA} & \multirow{2}[2]{*}{LC} & \multirow{2}[2]{*}{LL} & \multicolumn{3}{c}{Image-level} & \multicolumn{4}{c}{Pixel-level} \\
\cmidrule(r){5-7} \cmidrule(l){8-11} 
& & & & \multicolumn{1}{c}{AUROC} & \multicolumn{1}{c}{AP} & \multicolumn{1}{c}{$F_1$-max} & \multicolumn{1}{c}{AUROC} & \multicolumn{1}{c}{AP} & \multicolumn{1}{c}{$F_1$-max} & \multicolumn{1}{c}{AUPRO} \\ \midrule
   &    &    &    & 98.41 & 99.09 & 97.41  & 97.18 & 62.96 & 63.82  & 92.95 \\
\checkmark  &    &    &    & 99.06 & 99.54 & 98.31  & 97.62 & 66.22 & 66.70  & 93.71 \\
   & \checkmark  &    &    & 98.54 & 99.21 & 97.62  & 97.20 & 62.94 & 63.73  & 93.09 \\
   &    & \checkmark  &    & 98.35 & 99.04 & 97.43  & 97.10 & 61.05 & 62.73  & 92.60 \\
   &    &    & \checkmark  & 99.03 & 99.45 & 98.19  & 97.62 & 64.10 & 64.96  & 93.34 \\
\checkmark & \checkmark  &    &    & 99.27 & 99.62 & 98.63  & 97.85 & 67.36 & 67.33  & 94.16 \\
\checkmark &   & \checkmark   &    & 99.50 & 99.72 & 98.87  & 98.14 & 68.16 & 68.24  & 94.23 \\

\checkmark  &    & \checkmark  & \checkmark  &  99.52  &  \underline{99.73} &  98.92 &  \underline{98.20}  &   \underline{68.25}  &     \underline{68.34}   &  94.17   \\
\checkmark  & \checkmark  & \checkmark  &    & \underline{99.57} & \textbf{99.78} & \underline{99.00}  & \underline{98.20} & 67.93 & 68.21  & \underline{94.50} \\
\checkmark  & \checkmark  & \checkmark  & \checkmark  & \textbf{99.60} & \textbf{99.78} & \textbf{99.04}  & \textbf{98.35} & \textbf{69.29} & \textbf{69.17}  & \textbf{94.79} \\ \bottomrule
\end{tabular}
}
\label{tab3}
\end{table*}

\textbf{Implementation Details.}
 ViT-Base/14 (patchsize=14) pre-trained by DINOv2-R \cite{darcet2023vision} is used as the encoder by default. The drop rate of Noisy Bottleneck is 0.2 by default and increases to 0.4 on the diverse Real-IAD. Loose constraint with 2 groups is employed, and the anomaly map is given by the mean per-point cosine distance of the 2 groups. The input image is first resized to $448^2$ and then center-cropped to $392^2$, so the feature map ($28^2$) is large enough for anomaly localization. StableAdamW optimizer \cite{wortsman2023stable} with AMSGrad \cite{reddi2019convergence}  (more stable than AdamW \cite{loshchilov2017decoupled} in training) is utilized with $lr$=2e-3, $\beta$=(0.9,0.999) and $wd$=1e-4. The network is trained for 10,000 iterations (steps) on MVTec-AD and VisA, and 50,000 iterations on Real-IAD. Detailed settings are available in Appendix~\ref{sec:imp}.
 \subsection{Comparison to Multi-Class UAD SoTAs}

We compare the proposed Dinomaly with the most advanced UAD and MUAD methods \cite{deng2022anomaly,liu2023simplenet,zhang2023destseg,guo2023recontrast,zhang2023exploring,he2024diffusion,he2024mambaad}. Experimental results are presented in Table~\ref{tab1}, where Dinomaly surpasses compared methods by a large margin on all datasets and all metrics. On the most widely used MVTec-AD, Dinomaly produces image-level performance of \textbf{99.6}/\textbf{99.8}/\textbf{99.0} (\%) and pixel-level performance of \textbf{98.4}/\textbf{69.3}/\textbf{69.2}/\textbf{94.8}, outperforming previous SoTAs by \textbf{\textit{1.0}}/\textbf{\textit{0.2}}/\textbf{\textit{1.2}} and \textbf{\textit{0.7}}/\textbf{\textit{9.1}}/\textbf{\textit{7.7}}/\textbf{\textit{1.6}}. This result declares that the image-level performance on the MVTec-AD dataset is nearly saturated under the MUAD setting. On the popular VisA, Dinomaly achieves image-level performance of \textbf{98.7}/\textbf{98.9}/\textbf{96.2} and pixel-level performance of \textbf{98.7}/\textbf{53.2}/\textbf{55.7}/\textbf{94.5}, outperforming previous SoTAs by \textbf{\textit{3.2}}/\textbf{\textit{2.5}}/\textbf{\textit{4.2}} and \textbf{\textit{0.2}}/\textbf{\textit{5.3}}/\textbf{\textit{5.1}}/\textbf{\textit{2.6}}. On the Real-IAD that contains 30 classes, each with 5 camera views, we produce image-level and pixel-level performance of \textbf{89.3}/\textbf{86.8}/\textbf{80.2} and \textbf{98.8}/\textbf{42.8}/\textbf{47.1}/\textbf{93.9}, outperforming previous SoTAs by \textbf{\textit{3.0}}/\textbf{\textit{2.2}}/\textbf{\textit{3.2}} and \textbf{\textit{0.3}}/\textbf{\textit{4.9}}/\textbf{\textit{5.4}}/\textbf{\textit{3.4}}, indicating our scalability to extremely complex scenarios. Per-class performances and qualitative visualization are presented in Appendix \ref{sec:perclass} and \ref{sec:visual}. We also produce superior results on other popular UAD benchmarks, i.e., MPDD~\cite{jezek2021deep}, BTAD~\cite{mishra2021vt}, and Uni-Medical~\cite{zhang2024ader}, with I-AUROC of 97.2, 95.4, and 84.9, respectively, as shown in Table~\ref{tab:mpdd} in Appendix.

\subsection{Comparison to Class-Separated UAD SoTAs}
Dinomaly is also compared with class-separated SoTAs, as shown in Table~\ref{tab2}. Dinomaly under MUAD setting is comparable to conventional methods \cite{deng2022anomaly,liu2023simplenet,roth2022towards} that build individual models for each class. On MVTec-AD and VisA, multi-class Dinomaly (first row) is subjected to nearly no performance drop compared to its class-separated counterpart (second row). On the complicated Real-IAD that involves more classes and views, multi-class Dinomaly suffers a moderate performance drop but is still comparable to class-separated SoTAs.

\begin{table*}[!t]
  \centering
  \scriptsize
  \caption{Scaling of ViT model sizes on MVTec-AD (\%). Im/s (Troughoutput, image per second) is measured on NVIDIA RTX3090 with batch size=16. Results on VisA and Real-IAD are shown in Table~\ref{tab:archvisa}. \dag:default.}
   \resizebox{0.9\linewidth}{!}{
    \begin{tabular}{lcccccccccc}
    \toprule
    \multirow{2}[2]{*}{Arch.} & \multirow{2}[2]{*}{Params} & \multirow{2}[2]{*}{MACs} & \multirow{2}[2]{*}{Im/s} & \multicolumn{3}{c}{Image-level} & \multicolumn{4}{c}{Pixel-level} \\
\cmidrule(r){5-7} \cmidrule(l){8-11} 
& & & & \multicolumn{1}{c}{AUROC} & \multicolumn{1}{c}{AP} & \multicolumn{1}{c}{$F_1$-max} & \multicolumn{1}{c}{AUROC} & \multicolumn{1}{c}{AP} & \multicolumn{1}{c}{$F_1$-max} & \multicolumn{1}{c}{AUPRO} \\ \midrule
ViT-Small & 37.4M & 26.3G & 153.6  & 99.26 & 99.67 & 98.72  & 98.07 & 68.29 & 67.78  & 94.36 \\
ViT-Base\dag  & 148.0M & 104.7G & 58.1  & \underline{99.60} & \underline{99.78} & \underline{99.04}  & \underline{98.35} & \underline{69.29} & \underline{69.17}  & \underline{94.79} \\
ViT-Large & 275.3M & 413.5G & 24.2 & \textbf{99.77} & \textbf{99.92} & \textbf{99.45}  & \textbf{98.54} & \textbf{70.53} & \textbf{70.04}  & \textbf{95.09} \\

\bottomrule
\end{tabular}
}
\label{tab:arch}
\end{table*}


\begin{table}[!t]
\centering
\small
  \caption{Scaling input size on MVTec-AD (\%).  \dag: default. Compared methods yield degradation when increasing input size.}
\resizebox{0.95\linewidth}{!}{
\begin{tabular}{@{}llcc@{}}
\toprule
Method                      & Input Size & Image-Level       & Pixel-Level       \\ \midrule
\multirow{3}{*}{RD4AD}      & $256^2$\dag      & \textbf{94.6}/96.5/\textbf{96.1} & \textbf{96.1}/48.6/53.8/\textbf{91.1} \\
                            & $320^2$        & 93.2/\textbf{96.9}/95.6 & 95.7/\textbf{55.1}/\textbf{57.5}/\textbf{91.1} \\
                            & $384^2$        & 91.9/96.2/95.0   & 94.9/52.1/55.3/90.8          \\ \midrule
\multirow{3}{*}{ReContrast} & $256^2$\dag   & \textbf{98.3}/\textbf{99.4}/\textbf{97.6} & \textbf{97.1}/60.2/61.5/93.2     \\
                            & $320^2$      & 98.2/99.2/97.5	& 96.8/\textbf{61.8}/\textbf{62.6}/\textbf{93.3} \\
                            & $384^2$       & 95.2/98.0/96.4	&96.5/57.7/59.5/92.6 \\ \midrule
\multirow{3}{*}{Dinomaly}   & $280^2$        & \textbf{99.6}/\textbf{99.8}/\textbf{99.3} & 98.2/65.2/66.3/93.6          \\
                            & $336^2$        & \textbf{99.6}/\textbf{99.8}/99.2 & 98.3/67.2/67.8/94.2          \\
                            & $392^2$\dag        & \textbf{99.6}/\textbf{99.8}/99.0 & \textbf{98.4}/\textbf{69.3}/\textbf{69.2}/\textbf{94.8} \\ \bottomrule
\end{tabular}
}
\label{tab:inputsize}
\end{table}

\subsection{Ablation Study}
\label{sec:ab}
\textbf{Overall Ablation.} We conduct experiments to verify the effectiveness of the proposed elements, i.e., Noisy Bottleneck (NB), Linear Attention (LA), Loose Constraint (LC), and Loose Loss (LL). The already-powerful baseline (first row) is Dinomaly with noiseless MLP bottleneck, Softmax Attention, dense layer-to-layer supervision, and global cosine loss. This baseline is very similar to ViTAD~\cite{zhang2020exploring} and the ViT version of RD4AD~\cite{deng2022anomaly}. Results on MVTec-AD and VisA are shown in Table~\ref{tab3} and Table~\ref{tab:abvisa}, respectively. NB and LL can directly contribute to the model performance. LA and LC boost the performance with the presence of NB. The use of LC is not solely beneficial because LC makes the reconstruction too easy without injected noise.

\textbf{Model Scalability.} Previous works \cite{you2022unified,deng2022anomaly,zhang2023exploring} reported that anomaly detection networks do not follow the "scaling law". For example, RD4AD \cite{deng2022anomaly} found WideResNet50 better than WideResNet101 as the encoder backbone. ViTAD \cite{zhang2023exploring} found ViT-Small better than ViT-Base. On the contrary, as shown in Table~\ref{tab:arch}, the performance of the proposed Dinomaly benefits from scaling. Dinomaly equipped with ViT-Small has already produced state-of-the-art results. ViT-Large further boosts Dinomaly to an unprecedented higher record. This scalability enables users to choose an appropriate model size based on the computational resources available in their specific scenario. A comparison of computational costs with other methods is presented in Table~\ref{tab:cost}. In addition, training schedule can also be scaled up for even better performance without increasing inference costs, as demonstrated in Figure \ref{tab1} (\textcolor{gray}{Dinomaly$\uparrow$}).

\textbf{Input Scalability.} Though it seems unfair to compare Dinomaly with previous works that take smaller images as input, we contend that increasing their input size not only fails to benefit but actively undermines their performance, especially for image-level detection performance, as shown in Table~\ref{tab:inputsize}. Therefore, we follow the common comparison strategy based on "optimum vs. optimum". On the contrary, Dinomaly enjoys scaling input size for anomaly localization, while still producing SoTA performance given smaller images. Details are presented in Table~\ref{tab:resolution} in Appendix.

\begin{figure}[!t]
\centering
\centerline{\includegraphics[width=1\linewidth]{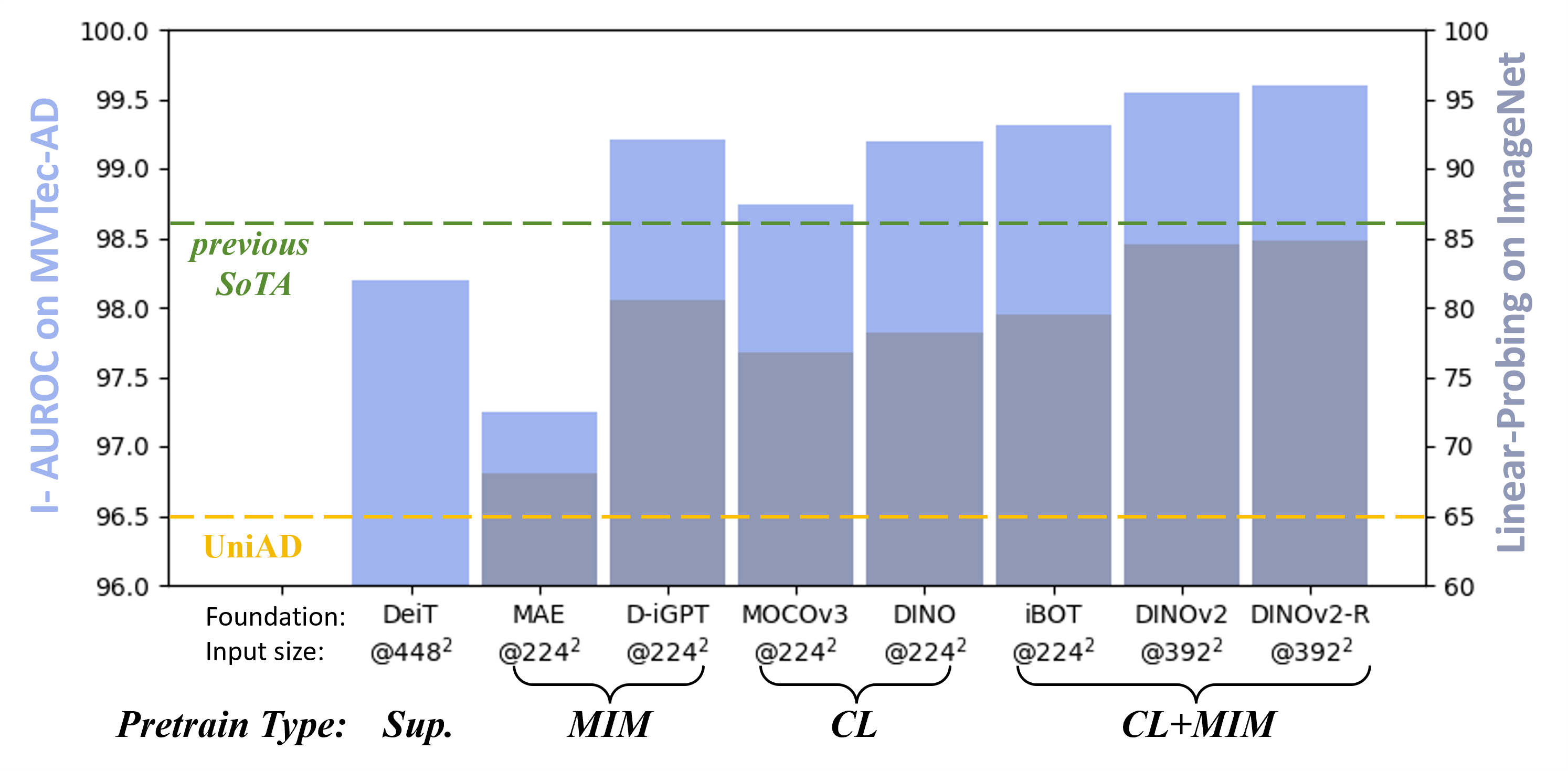}}
\caption{Image-level AUROC of Dinomaly equipped with various ViT foundations, and their linear-probing accuracy on ImageNet. MIM: Masked Image Modeling. CL: Contrastive Learning.}
\label{fig:backbone}
\end{figure}   

\textbf{ViT Foundations.}  We conduct extensive experiments to investigate the impact of diverse pre-trained ViT foundations, including DeiT \cite{touvron2021training}, MAE~\cite{he2022masked}, D-iGPT~\cite{ren2023rejuvenating}, MOCOv3~\cite{chen2021empirical}, DINO~\cite{caron2021emerging}, iBot~\cite{zhou2021ibot}, DINOv2~\cite{oquab2023dinov2}, and DINOv2-R \cite{darcet2023vision}.  As shown in Figure~\ref{fig:backbone}, Dinomaly is robust to the choice of backbone. Almost all foundation models can produce SoTA-level results with image-level AUROC higher than 98\%.  The only notable exception is MAE, which, without fine-tuning, was reported to be less effective across various unsupervised tasks, e.g, kNN and linear-probing \cite{oquab2023dinov2}. The optimal input size varies because the these backbones are pre-trained on different resolutions.  Interestingly, we found the anomaly detection performance to be strongly correlated with the accuracy of ImageNet linear-probing (freeze backbone \& only tune linear classifier) of the foundation model, suggesting the possibility of further improvement by simply adopting a future foundation model. Detailed results and analysis are presented in Appendix and Table~\ref{tab:pretrain}.

Additional experiments and results are detailed in the Appendix~\ref{sec:ablation}, encompassing evaluations of various pre-trained foundations, ablation studies of each components, hyperparameter optimization, and other in-depth analyses.

\section{Conclusion}
Dinomaly, a minimalistic UAD framework, is proposed to address the under-performed MUAD models in this paper. We present four key elements in Dinomaly, i.e., Foundation Transformer, Noisy MLP Bottleneck, Linear Attention, and Loose Reconstruction, that can boost the performance under the challenging MUAD setting without fancy modules and tricks. Extensive experiments on MVTec AD, VisA, and Real-IAD demonstrate our superiority over previous model-unified multi-class models and even recent class-separated models, indicating the feasibility of implementing a unified model in complicated scenarios free of severe performance degradation.

%% file: sec/X_suppl.tex
\clearpage
\setcounter{page}{1}
\maketitlesupplementary

\setcounter{table}{0}
\renewcommand{\thetable}{A\arabic{table}}

\setcounter{figure}{0}
\renewcommand{\thefigure}{A\arabic{figure}}

\setcounter{section}{0} 
\renewcommand{\thesection}{\Alph{section}} 

\begin{table*}[!h]
  \centering
  \tiny
  \caption{Comparison between pre-trained ViT foundations, conducted on MVTec-AD (\%). All models are ViT-Base. The patch size of DINOv2 and DINOv2-R is $14^2$; others are $16^2$. R$448^2$-C$392^2$ represents first resizing images to 448$\times$448, then center cropping to 392$\times$392.}
   \resizebox{\linewidth}{!}{
    \begin{tabular}{lccccccccc}
    \toprule
    \multirow{2}[2]{*}{\makecell[c]{Pre-Train\\Backbone}} & \multirow{2}[2]{*}{Type} & \multirow{2}[2]{*}{\makecell[c]{Image\\Size}} & \multicolumn{3}{c}{Image-level} & \multicolumn{4}{c}{Pixel-level} \\
\cmidrule(r){4-6} \cmidrule(l){7-10} 
 & & & \multicolumn{1}{c}{AUROC} & \multicolumn{1}{c}{AP} & \multicolumn{1}{c}{$F_1$-max} & \multicolumn{1}{c}{AUROC} & \multicolumn{1}{c}{AP} & \multicolumn{1}{c}{$F_1$-max} & \multicolumn{1}{c}{AUPRO} \\ \midrule
DeiT\cite{touvron2021training} & Supervised & R$512^2$-C$448^2$   & 98.19 & 99.24 & 97.64  & 97.93 & 68.98 & 67.91  & 91.45 \\
MAE\cite{he2022masked} & MIM & R$512^2$-C$448^2$  & 96.27 & 98.33 & 95.44  & 96.96 & 62.89 & 63.32  & 89.85 \\
D-iGPT\cite{ren2023rejuvenating} & MIM & R$512^2$-C$448^2$  & 98.75 & 99.24 & 97.70  & 98.30 & 65.77 & 66.16  & 92.34 \\
MOCOv3\cite{chen2021empirical} & CL & R$512^2$-C$448^2$  & 98.47 & 99.42 & 97.36  & 98.52 & 70.99 & 69.41  & 92.83 \\
DINO\cite{caron2021emerging} & CL & R$512^2$-C$448^2$  & 98.97 & 99.58 & 98.14  & 98.52 & 70.89 & 69.02  & 93.48 \\
iBOT\cite{zhou2021ibot} & CL+MIM & R$512^2$-C$448^2$  & 99.22 & 99.67 & 98.57  & 98.60 & 70.78 & 69.92  & 93.33 \\
DINOv2\cite{oquab2023dinov2} & CL+MIM & R$448^2$-C$392^2$  & 99.55 & 99.81 & 99.13  & 98.26 & 68.35 & 68.79  & 94.83 \\
DINOv2-R\cite{darcet2023vision} & CL+MIM & R$448^2$-C$392^2$  & 99.60 & 99.78 & 99.04  & 98.35 & 69.29 & 69.17  & 94.79 \\
\midrule
DeiT\cite{touvron2021training} & Supervised & R$256^2$-C$224^2$   & 97.65 & 99.05 & 97.40  & 97.80 & 62.58 & 63.39  & 89.98 \\
MAE\cite{he2022masked} & MIM & R$256^2$-C$224^2$  & 97.25 & 98.84 & 96.94  & 97.78 & 63.00 & 64.01  & 90.95 \\
BEiTv2\cite{peng2002beitv2} & MIM & R$256^2$-C$224^2$  & 97.70 & 99.11 & 97.39  & 97.61 & 59.79 & 62.53  & 90.10 \\
D-iGPT\cite{ren2023rejuvenating} & MIM & R$256^2$-C$224^2$  & 99.21 & 99.66 & 98.47  & 98.08 & 60.05 & 63.05  & 91.78 \\
MOCOv3\cite{chen2021empirical} & CL & R$256^2$-C$224^2$  & 98.74 & 99.56 & 98.33  & 98.05 & 63.36 & 64.38  & 91.13 \\
DINO\cite{caron2021emerging} & CL & R$256^2$-C$224^2$  & 99.20 & 99.72 & 98.77  & 98.16 & 64.16 & 65.07  & 92.02 \\
iBOT\cite{zhou2021ibot} & CL+MIM & R$256^2$-C$224^2$  & 99.31 & 99.74 & 98.77  & 98.25 & 64.01 & 65.37  & 91.68 \\
DINOv2\cite{oquab2023dinov2} & CL+MIM & R$256^2$-C$224^2$  & 99.26 & 99.70 & 98.60  & 97.95 & 62.27 & 64.39  & 92.80 \\
DINOv2-R\cite{darcet2023vision} & CL+MIM & R$256^2$-C$224^2$  & 99.34 & 99.73 & 99.03  & 98.09 & 63.04 & 64.48  & 92.59 \\ 
\bottomrule
\end{tabular}
}
\label{tab:pretrain}
\end{table*}

\begin{table*}[!h]
  \centering
  \tiny
  \caption{Ablations of input size, conducted on MVTec-AD (\%). R$448^2$-C$392^2$ represents first resizing images to 448$\times$448, then center cropping to 392$\times$392.}
   \resizebox{0.9\linewidth}{!}{
    \begin{tabular}{lcccccccc}
    \toprule
    \multirow{2}[2]{*}{Image Size} & \multirow{2}[2]{*}{MACs} & \multicolumn{3}{c}{Image-level} & \multicolumn{4}{c}{Pixel-level} \\
\cmidrule(r){3-5} \cmidrule(l){6-9} 
 & & \multicolumn{1}{c}{AUROC} & \multicolumn{1}{c}{AP} & \multicolumn{1}{c}{$F_1$-max} & \multicolumn{1}{c}{AUROC} & \multicolumn{1}{c}{AP} & \multicolumn{1}{c}{$F_1$-max} & \multicolumn{1}{c}{AUPRO} \\ \midrule
 R$512^2$-C$448^2$ &  136.4G  & \textbf{99.67} & 99.81 & 99.12  & 98.33 & \underline{69.24} & \textbf{69.47}  & 94.76 \\
R$448^2$ &  136.4G  & 99.59 & 99.77 & 99.19  & \textbf{98.57} & 68.09 & 68.58  & \textbf{95.60} \\
R$448^2$-C$392^2$\dag &  104.7G  & 99.60 & 99.78 & 99.04  & 98.35 & \textbf{69.29} & \textbf{69.17}  & 94.79 \\
R$392^2$ & 104.7G  & 99.48 & 99.74 & 99.04  & 98.47 &  67.02 & 67.86  & \underline{95.34} \\
R$384^2$-C$336^2$ & 77.1G  & 99.61 & 99.78 & 99.22  & 98.27 & 67.22 &  67.77  & 94.24 \\
R$336^2$ & 77.1G  & \underline{99.63} & \textbf{99.84} & \underline{99.23}  & \underline{98.48} & 65.46 & 66.60  & 95.10 \\
R$320^2$-C$280^2$ & 53.7G  & 99.62 & \underline{99.81} & 99.07  & 98.21 & 65.21 & 66.34  & 93.57 \\
R$280^2$  & 53.7G & 99.46 & 99.75 & \textbf{99.27}  & 98.40 & 63.28 & 64.79  & 94.47 \\
\bottomrule
\end{tabular}
}
\label{tab:resolution}
\end{table*}

\section{Additional Related Work}
\label{sec:relatedwork}
Here, we discussed general methods for unsupervised anomlay detection.
\textit{Epistemic methods} are based on the assumption that the networks respond differently during inference between seen input and unseen input. Within this paradigm, \textit{pixel reconstruction} methods assume that the networks trained on normal images can reconstruct anomaly-free regions well, but poorly for anomalous regions. Auto-encoder (AE) \cite{bergmann2018improving,zavrtanik2021reconstruction}, variational auto-encoder (VAE) \cite{liu2020towards,dehaene2020anomaly}, or generative adversarial network (GAN) \cite{schlegl2019f,akcay2019ganomaly} are used to restore normal pixels. However, \textit{pixel reconstruction} models may also succeed in restoring unseen anomalous regions if they resemble normal regions in pixel values or the anomalies are barely noticeable \cite{deng2022anomaly}. Therefore, \textit{feature reconstruction} is proposed to construct features of pre-trained encoders instead of raw pixels \cite{deng2022anomaly,you2022unified,yang2020dfr}. To prevent the whole network from converging to a trivial solution, the parameters of the encoders are frozen during training. In \textit{feature distillation} \cite{salehi2021multiresolution,wang2021student}, the student network is trained from scratch to mimic the output features of the pre-trained teacher network with the same input of normal images, also based on the similar hypothesis that the student trained on normal samples only succeed in mimicking features of normal regions.

 \textit{Pseudo-anomaly} methods generate handcrafted defects on normal images to imitate anomalies, converting UAD to supervised classification \cite{li2021cutpaste} or segmentation tasks \cite{zavrtanik2021draem}. Specifically, CutPaste \cite{li2021cutpaste} simulates anomalous regions by randomly pasting cropped patches of normal images. DRAEM \cite{zavrtanik2021draem} constructs abnormal regions using Perlin noise as the mask and another image as the additive anomaly. DeTSeg ~\cite{zhang2023destseg} employs a similar anomaly generation strategy and combines it with feature reconstruction. SimpleNet \cite{liu2023simplenet} introduces anomaly by injecting Gaussian noise in the pre-trained feature space. These methods deeply rely on how well the pseudo anomalies match the real anomalies, which makes it hard to generalize to different datasets. 
 
\textit{Feature statistics} methods \cite{defard2021padim,roth2022towards,sheynin2021hierarchical,lee2022cfa} memorize all  normal features (or their modeled distribution) extracted by networks pre-trained on large-scale datasets and match them with test samples during inference. Since these methods require memorizing, processing, and matching nearly all features from training samples, they are computationally expensive in both training and inference, especially when the training set is large.

\textbf{Scope of Application.} In this work, we focus on \textbf{sensory AD} that detects regional or structural anomalies (common in practical applications such as industrial inspection, medical disease screening, etc.), which is distinguished from \textbf{semantic AD}. In sensory AD, normal and anomalous samples are the same objects except for anomaly, e.g. good cable vs. spoiled cable. In semantic AD, the class of normal samples and anomalous samples are semantically different, e.g. animals vs. vehicles. Semantic AD methods usually utilize and compare the global representation of images, which generally do not suffer from the issues of multi-class setting discussed in this paper..

\section{Full Implementation Details}
\label{sec:imp}

 ViT-Base/14 (patch size=14) pre-trained by DINOv2 with registers (DINOv2-R) \cite{darcet2023vision} is utilized as the encoder by default. The discard rate of Dropout in Noisy Bottleneck is 0.2 by default, which is increased to 0.4 for the diverse Real-IAD. Loose constraint with 2 groups and $\mathcal{L}_{global-hm}$ loss are used by default. The input image is first resized to $448^2$ and then center-cropped to $392^2$, so that the feature map ($28^2$) is large enough for localization. As previously discussed, the middle 8 layers of 12-layer ViT-Base are used for reconstruction and feeding the bottleneck. ViT-Small also has 12 layers, which is the same. ViT-Large contains 24 layers; therefore, we use the [4,6,8,...18] layers (index start from 0). The decoder always contains 8 layer.
 
 StableAdamW optimizer \cite{wortsman2023stable} with AMSGrad \cite{reddi2019convergence} is utilized with $lr$ (learning rate)=2e-3, $\beta$=(0.9,0.999), $wd$ (weight decay)=1e-4 and $eps$=1e-10. The network is trained for 10,000 iterations for MVTec-AD and VisA and 50,000 iterations for Real-IAD under MUAD setting. The network is trained for 5,000 iterations on each class under the class-separated UAD setting. The $lr$ warms up from 0 to 2e-3 in the first 100 iterations and cosine anneals to 2e-4 throughout the training. The discarding rate in Equation \ref{eq:ghm} linearly rises from 0\% to 90\% in the first 1,000 iterations as warm-up (500 iters for class-separated setting). The anomaly map is obtained by upsampling the point-wise cosine distance between encoder and decoder feature maps (averaging if more than one pair or group). The mean of the top 1\% pixels in an anomaly map is used as the image anomaly score. All experiments are conducted with random seed=1 with cuda deterministic for invariable weight initialization and batch order. Codes are implemented with Python 3.8 and PyTorch 1.12.0 cuda 11.3, and run on NVIDIA GeForce RTX3090 GPUs (24GB).

Most results of compared MUAD SoTAs are directly drawn from a benchmark paper ADer \cite{zhang2024ader}. We express great thanks for their wonderful work.

\begin{table*}[t]
  \centering
  \tiny
  \caption{Scaling of ViT architectures on VisA and Real-IAD (\%). \dag: default.}
   \resizebox{0.9\textwidth}{!}{
    \begin{tabular}{ccccccccc}
    \toprule
    \multirow{2}[2]{*}{Dateset} & \multirow{2}[2]{*}{Arch.} & \multicolumn{3}{c}{Image-level} & \multicolumn{4}{c}{Pixel-level} \\
\cmidrule(r){3-5} \cmidrule(l){6-9} 
& &\multicolumn{1}{c}{AUROC} & \multicolumn{1}{c}{AP} & \multicolumn{1}{c}{$F_1$-max} & \multicolumn{1}{c}{AUROC} & \multicolumn{1}{c}{AP} & \multicolumn{1}{c}{$F_1$-max} & \multicolumn{1}{c}{AUPRO} \\
\hline
    \multirow{3}[0]{*}{VisA~\cite{zou2022spot}} & ViT-Small & 97.94  & 98.09  & 95.33  & 98.57  & 51.19  & 55.10  & 93.71 \\
          & ViT-Base\dag & \underline{98.73} & \underline{98.87} & \textbf{96.18} & \underline{98.74} & \underline{53.23} & \underline{55.69} & \underline{94.50} \\
          & ViT-Large & \textbf{98.85} & \textbf{99.09} & \underline{96.12} & \textbf{99.10} & \textbf{55.68} & \textbf{57.33} & \textbf{94.76} \\
\hline
    \multirow{3}[0]{*}{Real-IAD~\cite{wang2024real}} & ViT-Small & 89.10  & 86.91  & 79.87  & 98.69  & 41.88  & 46.74  & \underline{94.08} \\
          & ViT-Base\dag & \underline{89.33} & \underline{86.77} & \underline{80.17} & \underline{98.84} & \underline{42.79} & \underline{47.10} & 93.86 \\
          & ViT-Large & \textbf{90.07} & \textbf{87.57} & \textbf{80.90} & \textbf{99.02} & \textbf{44.29} & \textbf{48.36} & \textbf{94.37} \\
          \bottomrule
\end{tabular}
}
\label{tab:archvisa}
\end{table*}

 \begin{table*}[!h]
  \centering
    \tiny
  \caption{Ablations of Dinomaly elements on VisA (\%). NB: Noisy Bottleneck. LA: Linear Attention. LC: Loosen Constraint (2 groups). LL: Loosen Loss.}
   \resizebox{0.9\linewidth}{!}{
    \begin{tabular}{ccccccccccc}
    \toprule
    \multirow{2}[2]{*}{NB} & \multirow{2}[2]{*}{LA} & \multirow{2}[2]{*}{LC} & \multirow{2}[2]{*}{LL} & \multicolumn{3}{c}{Image-level} & \multicolumn{4}{c}{Pixel-level} \\
\cmidrule(r){5-7} \cmidrule(l){8-11} 
& & & & \multicolumn{1}{c}{AUROC} & \multicolumn{1}{c}{AP} & \multicolumn{1}{c}{$F_1$-max} & \multicolumn{1}{c}{AUROC} & \multicolumn{1}{c}{AP} & \multicolumn{1}{c}{$F_1$-max} & \multicolumn{1}{c}{AUPRO} \\ \midrule
   &    &    &    & 95.81 & 96.35 & 92.06  & 97.97 & 47.88 & 52.55  & 93.43 \\
\checkmark  &    &    &    & 97.38 & 97.74 & 94.07  & 97.84 & 50.42 & 54.57  & 93.71 \\
   & \checkmark  &    &    & 95.74 & 96.23 & 91.87  & 98.01 & 47.89 & 52.58  & 93.34 \\
   &    & \checkmark  &    & 96.39 & 97.01 & 92.54  & 97.37 & 46.80 & 51.66  & 92.75 \\
   &    &    & \checkmark  & 96.93 & 97.26 & 93.32  & 98.37 & 49.52 & 53.59  & 94.11 \\
\checkmark & \checkmark  &    &    & 97.52 & 97.75 & 94.33  & 98.06 & 51.49 & 55.09  & 93.75 \\
\checkmark &   & \checkmark   &    & 98.06 & 98.37 & 95.18  & 98.21 & 51.43 & 54.89  & 93.94 \\

\checkmark  &    & \checkmark  & \checkmark  &  \underline{98.57}  &  \underline{98.77} &  \underline{95.75} &  \underline{98.57}  & 52.29  &   55.38   &  \underline{94.28}   \\
\checkmark  & \checkmark  & \checkmark  &    & 98.22 & 98.43 & 95.27  & 98.51 & \underline{53.11} & \underline{55.48}  & 94.24 \\
\checkmark  & \checkmark  & \checkmark  & \checkmark  & \textbf{98.73} & \textbf{98.87} & \textbf{96.18} & \textbf{98.74} & \textbf{53.23} & \textbf{55.69} & \textbf{94.50} \\ \bottomrule
\end{tabular}
}
\label{tab:abvisa}
\end{table*}

 \section{Additional Ablation and Experiment}
 \label{sec:ablation}

\textbf{Pre-Trained Foundations.} The representation quality of the frozen backbone Transformer is of great significance to unsupervised anomaly detection. We conduct extensive experiments to probe the impact of different pre-training methods, including supervised learning and self-supervised learning. DeiT \cite{touvron2021training} is trained on ImageNet\cite{deng2009imagenet} in a supervised manner by distilling CNNs. MAE~\cite{he2022masked}, BEiTv2~\cite{peng2002beitv2}, and D-iGPT~\cite{ren2023rejuvenating} are based on masked image modeling (MIM). Given input images with masked patches, MAE~\cite{he2022masked} is optimized to restore raw pixels; BEiTv2~\cite{peng2002beitv2} is trained to predict the token index of VQ-GAN and CLIP; D-iGPT~\cite{ren2023rejuvenating} is trained to predict the features of CLIP model. MOCOv3~\cite{chen2021empirical} is based on contrastive learning (CL), pulling the representations of the similar images and pushing those of different images. DINO~\cite{caron2021emerging} is based on positive-pair contrastive learning, which is also referred to as self-distillation. It trains the network to produce similar feature representations given two views (augmentations) of the same image. iBot~\cite{zhou2021ibot} and DINOv2~\cite{oquab2023dinov2} combine MIM and CL strategies, marking the SoTA of self-supervised foundation models. DINOv2-R \cite{darcet2023vision} is a variation of DINOv2 that employs 4 extra register tokens.

It is noted that most models are pre-trained with the image resolution of $224\times224$, except that DINOv2~\cite{oquab2023dinov2} and DINOv2-R~\cite{darcet2023vision} have extra a high-resolution training phase with $518\times518$. Directly using the pre-trained weights on a different resolution for UAD without fine-tuning like other supervised tasks can cause generalization problems. Therefore, by default, we still keep the feature size of all compared models to $28\times28$, i.e., the input size is $392\times392$ for ViT-Base/14 and  $448\times448$ for ViT-Base/16. Additionally, we train Dinomaly with the low-resolution input size of $224\times224$. 

The results are presented in Table~\ref{tab:pretrain}. Within Dinomaly, nearly all foundation models can produce SoTA-level results with image-level AUROC higher than 98\%. Generally speaking, CL+MIM combined models outperform MIM and CL models. In addition, most foundations do not benefit from a higher resolution for image-level performance but suffer from it, indicating the lack of generalization on a input size different from pre-training; while as expected, DINOv2 and DINOv2-R pre-trained on larger inputs can better benefit from higher resolution in Dinomaly. Because some methods, i.e., D-iGPT, DINO, and iBOT, produce similar results to DINOv2 in $224\times224$, we expect that they also have the potential to be as powerful in Dinomaly if they are pre-trained in high-resolution.
Employing MAE produces the worst results. MAE was also tested as the backbone of ViTAD\cite{zhang2023exploring}, resulting in undesirable performances (I-AUROC=95.3), which was attributed to the weak semantic expression caused by the pretraining strategy. It is also noted that MAE is bad in other unsupervised tasks such as ImageNet kNN; therefore, MAE is considered to be less effective in tasks without finetuning.

\begin{table*}[!h]
  \centering
  \tiny
  \caption{Ablations of Dropout rates in Noisy Bottleneck, conducted on MVTec-AD (\%). \dag: default.}
   \resizebox{0.8\linewidth}{!}{
    \begin{tabular}{lccccccc}
    \toprule
    \multirow{2}[2]{*}{Dropout rate} & \multicolumn{3}{c}{Image-level} & \multicolumn{4}{c}{Pixel-level} \\
\cmidrule(r){2-4} \cmidrule(l){5-8} 
 & \multicolumn{1}{c}{AUROC} & \multicolumn{1}{c}{AP} & \multicolumn{1}{c}{$F_1$-max} & \multicolumn{1}{c}{AUROC} & \multicolumn{1}{c}{AP} & \multicolumn{1}{c}{$F_1$-max} & \multicolumn{1}{c}{AUPRO} \\ \midrule
0 (noiseless)    & 99.19 & 99.55 & 98.51  & 97.55 & 63.11 & 64.39  & 93.33 \\
0.1    & 99.54 & 99.75 & 98.90  & \textbf{98.35} & \textbf{69.46} & \textbf{69.19}  & 94.53 \\
0.2 \dag    & 99.60 & 99.78 & 99.04  & \textbf{98.35} & \underline{69.29} & \underline{69.17}  & \textbf{94.79} \\
0.3    & \textbf{99.65} & \textbf{99.83} & \underline{99.16} & \underline{98.34} & 68.46 & 68.81  & \underline{94.63} \\
0.4    & \underline{99.64} & \underline{99.80} & \textbf{99.23}  & 98.22 & 67.95 & 68.33  & 94.57 \\
0.5   & 99.56 & 99.81 & 99.14  & 98.15 & 67.43 & 67.82  & 94.64 \\
\bottomrule
\end{tabular}
}
\label{tab:noise}
\end{table*}

\begin{table*}[!h]
  \centering
  \tiny
  \caption{Ablations of reconstruction constraint, conduected on MVTec-AD (\%). \dag: default.}
   \resizebox{0.95\linewidth}{!}{
    \begin{tabular}{lccccccc}
    \toprule
    \multirow{2}[2]{*}{Constraints} & \multicolumn{3}{c}{Image-level} & \multicolumn{4}{c}{Pixel-level} \\
\cmidrule(r){2-4} \cmidrule(l){5-8} 
 & \multicolumn{1}{c}{AUROC} & \multicolumn{1}{c}{AP} & \multicolumn{1}{c}{$F_1$-max} & \multicolumn{1}{c}{AUROC} & \multicolumn{1}{c}{AP} & \multicolumn{1}{c}{$F_1$-max} & \multicolumn{1}{c}{AUPRO} \\ \midrule
layer-to-layer (dense, every 1)   & 99.39 & 99.68 & 98.73  & 98.12 & 68.55 & \underline{68.63}  & 94.28 \\
layer-to-layer (sparse, every 2)    & 99.52 & 99.73 & 98.95  & 98.16 & \underline{68.89} & 68.57  & \underline{94.40} \\
layer-to-layer (sparse, every 4)    & 99.54 & 99.77 & 99.05  & 98.04 & 66.69 & 67.17  & 94.07 \\
layer-to-cat-layer (every 2)    & 99.48 & 99.71 & \underline{99.26}  & 97.83 & 62.29 & 62.91  & 93.16 \\
group-to-group (1 group)    & \textbf{99.64} & \textbf{99.80} & \textbf{99.36}  & \underline{98.18} & 64.79 & 65.40  & 93.96 \\
group-to-group (2 groups)\dag   & \underline{99.60} & \underline{99.78} & 99.04  & \textbf{98.35} & \textbf{69.29} & \textbf{69.17}  & \textbf{94.79} \\
\bottomrule
\end{tabular}
}
\label{tab:scheme}
\end{table*}

\textbf{Input Size.} The patch size of ViTs (usually $14\times14$ or $16\times16$) is much larger than the stem layer's down-sampling rate of CNNs (usually $4\times4$), resulting in smaller feature map size. For dense prediction tasks like semantic segmentation, ViTs usually employ a large input image size \cite{oquab2023dinov2}. This practice holds in anomaly localization as well. In Table~\ref{tab:resolution}, we present the results of Dinomaly with different input resolutions. Following PatchCore \cite{roth2022towards},  by default, we adopt center-crop preprocessing to reduce the influence of background, which can also cause unreachable anomalies at the edge of images. Experimental results demonstrate our robustness to input size. While small image size is enough for image-level anomaly detection, larger inputs are beneficial to anomaly localization. All experiments evaluate localization performance in a unified size of $256\times256$ for fairness. 

\textbf{Scalability on VisA and Real-IAD.} We demonstrate the performance of different ViT sizes on VisA and Real-IAD in Table~\ref{tab:archvisa}.

\begin{table*}[!h]
  \centering
  \tiny
  \caption{Comparison between Convolutional block, Softmax Attention, and Linear Attention as the spatial mixer of decoder, conducted on MVTec-AD (\%).}
   \resizebox{0.9\linewidth}{!}{
    \begin{tabular}{lccccccc}
    \toprule
    \multirow{2}[2]{*}{Spatial Mixer} & \multicolumn{3}{c}{Image-level} & \multicolumn{4}{c}{Pixel-level} \\
\cmidrule(r){2-4} \cmidrule(l){5-8} 
 & \multicolumn{1}{c}{AUROC} & \multicolumn{1}{c}{AP} & \multicolumn{1}{c}{$F_1$-max} & \multicolumn{1}{c}{AUROC} & \multicolumn{1}{c}{AP} & \multicolumn{1}{c}{$F_1$-max} & \multicolumn{1}{c}{AUPRO} \\ \midrule
ConvBlock $3\times3$     & 99.45 & 99.63 & 98.64  & 98.05 & 65.35 & 68.07  & 94.17 \\
ConvBlock $5\times5$    & 99.41 & 99.62 & 98.86  & 97.99 & 66.64 & 67.47  & 94.24 \\
ConvBlock $7\times7$    & 99.42 & 99.65 & 98.86  & 98.01 & 67.57 & 67.94  & 94.45 \\
\midrule
Softmax Attention    & 99.52  &  99.73 &  98.92 & 98.20  &  68.25  & 68.34   &  94.17 \\
Softmax Attention w/ Neighbour-Mask  $n=1$    & 99.51  &  99.71 &  98.90 &  98.17  &  67.86  &   67.92   &  94.27 \\
 Softmax Attention w/ Neighbour-Mask $n=3$    & \underline{99.56}  &  99.76 &  \underline{99.05} &  98.28  &  69.26  &   68.17   &  94.50 \\
\midrule
Linear Attention   &  \textbf{99.60} & \underline{99.78} & 99.04  & \underline{98.35} & \underline{69.29} & \underline{69.17}  & \textbf{94.79} \\
Linear Attention w/ Neighbour-Mask  $n=1$  &  \textbf{99.60} & \underline{99.78} & 99.04  & 98.32 & 68.77 & 68.72  & \underline{94.75} \\
Linear Attention w/ Neighbour-Mask $n=3$  &  \textbf{99.60} & \textbf{99.80} & \textbf{99.14}  & \textbf{98.38} & \textbf{69.65} & \textbf{69.38}  & 94.70 \\
\bottomrule
\end{tabular}
}
\label{tab:conv}
\end{table*}

\begin{table*}[!h]
  \centering
  \tiny
  \caption{Dropout \textit{vs.} feature jitter, conducted on MVTec-AD (\%).}
   \resizebox{0.8\linewidth}{!}{
    \begin{tabular}{lccccccc}
    \toprule
    \multirow{2}[2]{*}{Noise type} & \multicolumn{3}{c}{Image-level} & \multicolumn{4}{c}{Pixel-level} \\
\cmidrule(r){2-4} \cmidrule(l){5-8} 
 & \multicolumn{1}{c}{AUROC} & \multicolumn{1}{c}{AP} & \multicolumn{1}{c}{$F_1$-max} & \multicolumn{1}{c}{AUROC} & \multicolumn{1}{c}{AP} & \multicolumn{1}{c}{$F_1$-max} & \multicolumn{1}{c}{AUPRO} \\ \midrule
No Noise    & 99.19 & 99.55 & 98.51  & 97.55 & 63.11 & 64.39  & 93.33 \\ \midrule

Patch Masking p=0.1    & 99.27 & 99.60 & 98.80  & 97.92 & 67.15 & 66.90  & 94.18 \\
Patch Masking p=0.2    & 99.17 & 99.56 & 98.59  & 97.75 & 66.55 & 66.32  & 94.11 \\
Patch Masking p=0.3     & 99.11 & 99.54 & 98.37  & 97.53 & 65.48 & 65.96  & 93.84 \\
Patch Masking p=0.4     & 99.20 & 99.59 & 98.53  & 97.71 & 65.58 & 66.36  & 94.15 \\ \midrule

Feature Jitter \textit{scale}=1    & 99.23 & 99.54 & 98.48  & 97.58 & 63.22 & 64.31  & 93.55 \\
Feature Jitter \textit{scale}=5    & 99.24 & 99.57 & 98.55  & 97.84 & 65.28 & 65.81  & 93.75 \\
Feature Jitter \textit{scale}=10    & 99.46 & 99.73 & 99.12  & 98.19 & 67.59 & 67.80  & 94.19 \\
Feature Jitter \textit{scale}=20    & 99.59 & 99.79 & 99.04  & 98.23 & 67.93 & 68.21  & 94.40 \\ \midrule

Dropout p=0.1    & 99.54 & 99.75 & 98.90  & \textbf{98.35} & \textbf{69.46} & \textbf{69.19}  & 94.53 \\
Dropout p=0.2   & 99.60 & 99.78 & 99.04  & \textbf{98.35} & \underline{69.29} & \underline{69.17}  & \textbf{94.79} \\
Dropout p=0.3    & \textbf{99.65} & \textbf{99.83} & \underline{99.16} & \underline{98.34} & 68.46 & 68.81  & \underline{94.63} \\
Dropout p=0.4    & \underline{99.64} & \underline{99.80} & \textbf{99.23}  & 98.22 & 67.95 & 68.33  & 94.57 \\
\bottomrule
\end{tabular}
}
\label{tab:dpfj}
\end{table*}

\begin{table*}[!h]
  \centering
  \tiny
  \caption{Integrating the essense of Noisy Bottleneck (NB) and Loose Loss (LL) on RD4AD, conducted on MVTec-AD (\%). \dag: Reproduction in our framework; ReLU in ResNet decoder is replaced by GELU, StableAdamW optimizer is used.}
   \resizebox{0.9\linewidth}{!}{
    \begin{tabular}{lccccccccc}
    \toprule
    \multirow{2}[2]{*}{Method} & \multirow{2}[2]{*}{NB} & \multirow{2}[2]{*}{LL} & \multicolumn{3}{c}{Image-level} & \multicolumn{4}{c}{Pixel-level} \\
\cmidrule(r){4-6} \cmidrule(l){7-10} 
 & & &\multicolumn{1}{c}{AUROC} & \multicolumn{1}{c}{AP} & \multicolumn{1}{c}{$F_1$-max} & \multicolumn{1}{c}{AUROC} & \multicolumn{1}{c}{AP} & \multicolumn{1}{c}{$F_1$-max} & \multicolumn{1}{c}{AUPRO} \\ \midrule
RD4AD\dag &  &   & 97.8	&99.1	&97.2	&96.4	& 58.0	&59.3	&91.9 \\
RD4AD &\checkmark &    & 98.4	&99.4	&97.9	&97.2	&58.6	&60.4	&92.9 \\
RD4AD &  &\checkmark   & 98.2	&99.2	&97.5	&96.8	&60.0	&61.1	&92.7 \\
RD4AD &\checkmark &\checkmark   & 98.5	&99.4	&97.8	&97.2	&59.6	&61.2	&93.0 \\
\bottomrule
\end{tabular}
}
\label{tab:cnn}
\end{table*}

\begin{table*}[!h]
  \centering
  \tiny
  \caption{Scaling properties of a previous ViT-based method, ViTAD\cite{zhang2023exploring} on MVTec-AD. \dag: their original setting.}
   \resizebox{0.9\linewidth}{!}{
    \begin{tabular}{lccccccccc}
    \toprule
    \multirow{2}[2]{*}{Method} & \multirow{2}[2]{*}{\makecell[c]{Pre-Train\\Backbone}} & \multirow{2}[2]{*}{\makecell[c]{Input\\Size}} & \multicolumn{3}{c}{Image-level} & \multicolumn{4}{c}{Pixel-level} \\
\cmidrule(r){4-6} \cmidrule(l){7-10} 
 & & &\multicolumn{1}{c}{AUROC} & \multicolumn{1}{c}{AP} & \multicolumn{1}{c}{$F_1$-max} & \multicolumn{1}{c}{AUROC} & \multicolumn{1}{c}{AP} & \multicolumn{1}{c}{$F_1$-max} & \multicolumn{1}{c}{AUPRO} \\ \midrule
ViTAD\dag & DINO & $256^2$  & 98.3	&99.4	&97.3&	97.7&	55.3&	58.7	&91.4 \\
ViTAD & MAE &  $256^2$  & 95.3	&97.7	&95.2	&97.4	&53.0	&56.2	&90.6 \\
ViTAD & DINOv2 & $256^2$   & 98.7	&99.4	&98.1	&97.6	&55.3	&59.1	&92.7 \\
ViTAD & DINOv2-R & $256^2$   & 98.5	&99.3	&97.8&	97.4	&54.5	&59.2	&92.8 \\ \midrule

ViTAD\dag & DINO & $256^2$   & 98.3	&99.4	&97.3&	97.7&	55.3&	58.7	&91.4 \\
ViTAD & DINO & $320^2$   & 98.3	&99.2	&97.1	&97.6	&61.3	&63.3	&92.4 \\
ViTAD & DINO & $384^2$  & 97.8	&98.9	&96.3	&97.5	&62.5	&63.7	&92.4 \\

\bottomrule
\end{tabular}
}
\label{tab:vitad}
\end{table*}

\begin{table*}[!h]
  \centering
  \tiny
  \caption{Matching previous methods in computation consumption. Dinomaly can be easily scaled by model size and input size.}
   \resizebox{\linewidth}{!}{
    \begin{tabular}{lcccccccc}
    \toprule
    \multirow{2}[2]{*}{Method} & \multirow{2}[2]{*}{Params} & \multirow{2}[2]{*}{MACs} & \multicolumn{3}{c}{MVTec-AD~\cite{bergmann2019mvtec}} & \multicolumn{3}{c}{VisA~\cite{zou2022spot}} \\
\cmidrule(r){4-6} \cmidrule(l){7-9} 
 & & & \multicolumn{1}{c}{I-AUROC} & \multicolumn{1}{c}{P-AUROC} & \multicolumn{1}{c}{P-AUPRO} & \multicolumn{1}{c}{I-AUROC} & \multicolumn{1}{c}{P-AUROC} & \multicolumn{1}{c}{P-AUPRO} \\ \midrule
  DiAD~\cite{he2024diffusion} & 1331M&	451.5G&	97.2&	96.8&	90.7&	86.8&	96.0&	75.2 \\
 ReContrast~\cite{guo2023recontrast} & 154.2M&	67.4G&	98.3&	97.1&	93.2&	95.5&	98.5&	91.9 \\
RD4AD~\cite{deng2022anomaly} & 126.7M & 32.1G   & 94.6 & 96.1 &	91.1&92.4	&98.1&	91.8 \\
ViTAD~\cite{zhang2023exploring} & 39.0M &	9.7G&	98.3&	97.7&	91.4&	90.5&	98.2&	85.1 \\
 \midrule
Dinomaly-Base-$392^2$ & 148M & 104.7G  & 99.6 & 98.4 & 94.8  & 98.7 & 98.7 & 94.5\\
Dinomaly-Base-$280^2$ & 148M & 53.7G  & 99.6 & 98.2 & 93.6  & 97.8 & 98.7 & 92.4  \\
Dinomaly-Small-$392^2$ & 37.4M	&26.2G	&99.3	&98.1	&94.4	&97.9	&98.6	&93.7 \\
Dinomaly-Small-$280^2$  &37.4M	&14.5G	&99.3	&98.0	&93.4	&96.5	&98.5	&90.9 \\
\bottomrule
\end{tabular}
}
\label{tab:cost}
\end{table*}

\begin{table*}[!h]
  \centering
  \tiny
  \caption{Results of 5 random seeds on MVTec-AD (\%).}
   \resizebox{\linewidth}{!}{
    \begin{tabular}{lccccccc}
    \toprule
    \multirow{2}[2]{*}{Random Seed} & \multicolumn{3}{c}{Image-level} & \multicolumn{4}{c}{Pixel-level} \\
\cmidrule(r){2-4} \cmidrule(l){5-8} 
 & \multicolumn{1}{c}{AUROC} & \multicolumn{1}{c}{AP} & \multicolumn{1}{c}{$F_1$-max} & \multicolumn{1}{c}{AUROC} & \multicolumn{1}{c}{AP} & \multicolumn{1}{c}{$F_1$-max} & \multicolumn{1}{c}{AUPRO} \\ \midrule
seed=1   &  99.60 & 99.78 & 99.04  & 98.35 & 69.29 & 69.17  & 94.79 \\
seed=2   &  99.63 & 99.79 & 99.12  & 98.33 & 68.73 & 68.91  & 94.63 \\
seed=3   &  99.63 & 99.79 & 99.16  & 98.31 & 68.70 & 68.93  & 94.60 \\
seed=4   &  99.56 & 99.74 & 99.02  & 98.33 & 69.04 & 69.09  & 94.70 \\
seed=5   &  99.59 & 99.77 & 99.02 & 98.32 & 68.64 & 68.47  & 94.51 \\
mean$\pm$std   & 99.60$\pm$0.03 & 99.77$\pm$0.02 & 99.07$\pm$0.06  & 98.33$\pm$0.01 & 68.88$\pm$0.25 & 68.91$\pm$0.24  & 94.65$\pm$0.09 \\
\bottomrule
\end{tabular}
}
\label{tab:seed}
\end{table*}

\begin{table*}[t]
  \centering
  \tiny
  \caption{Performance on MPDD and BTAD under \textbf{multi-class} UAD setting (\%). \dag: method designed for MUAD.}
   \resizebox{0.9\textwidth}{!}{
    \begin{tabular}{ccccccccc}
    \toprule
    \multirow{2}[2]{*}{Dateset} & \multirow{2}[2]{*}{Method} & \multicolumn{3}{c}{Image-level} & \multicolumn{4}{c}{Pixel-level} \\
\cmidrule(r){3-5} \cmidrule(l){6-9} 
& &\multicolumn{1}{c}{AUROC} & \multicolumn{1}{c}{AP} & \multicolumn{1}{c}{$F_1$-max} & \multicolumn{1}{c}{AUROC} & \multicolumn{1}{c}{AP} & \multicolumn{1}{c}{$F_1$-max} & \multicolumn{1}{c}{AUPRO} \\
\hline
    \multirow{8}[0]{*}{MPDD~\cite{jezek2021deep}} & RD4AD~\cite{deng2022anomaly} & 90.3  & 92.8  & 90.5  & \underline{98.3}  & 39.6  & 40.6  & 95.2 \\
          & SimpleNet~\cite{liu2023simplenet} & 90.6  & \underline{94.1}  & 89.7  & 97.1  & 33.6  & 35.7  & 90.0\\
          & DeSTSeg~\cite{zhang2023destseg} & \underline{92.6}  & 91.8  & \underline{92.8}  & 90.8  & 30.6  & 32.9  & 78.3\\
          & UniAD~\cite{you2022unified}\dag & 80.1  & 83.2  & 85.1  & 95.4  & 19.0  & 25.6  & 83.8\\
          & DiAD~\cite{he2024diffusion}\dag  & 85.8  & 89.2  & 86.5  & 91.4  & 15.3  & 19.2  & 66.1\\
          & ViTAD~\cite{zhang2023exploring}\dag & 87.4 & 90.8 & 87.0 & 97.8 & \underline{44.1} & \underline{46.4} & \underline{95.3} \\
          & MambaAD~\cite{he2024mambaad}\dag  & 89.2 & 93.1 & 90.3 & 97.7 & 33.5 & 38.6 & 92.8  \\
          & \textbf{Dinomaly} (Ours) &\textbf{97.2} & \textbf{98.4} & \textbf{96.0} & \textbf{99.1} & \textbf{59.5} & \textbf{59.4} & \textbf{96.6} \\
\hline
    \multirow{8}[0]{*}{BTAD~\cite{mishra2021vt}} & RD4AD~\cite{deng2022anomaly} & 94.1  & 96.8  & 93.8  & \textbf{98.0}  & 57.1  & \underline{58.0}  & \textbf{79.9} \\
          & SimpleNet~\cite{liu2023simplenet} & 94.0  & 97.9  & 93.9  & 96.2  & 41.0  & 43.7  & 69.6\\
          & DeSTSeg~\cite{zhang2023destseg} & 93.5  & 96.7  & 93.8  & 94.8  & 39.1  & 38.5  & 72.9\\
          & UniAD~\cite{you2022unified}\dag & \underline{94.5}  & \textbf{98.4}  & \underline{94.9}  & 97.4  & 52.4  & 55.5  & \underline{78.9}\\
          & DiAD~\cite{he2024diffusion}\dag  & 90.2  & 88.3  & 92.6  & 91.9  & 20.5  & 27.0  & 70.3\\
          & ViTAD~\cite{zhang2023exploring}\dag & 94.0 & 97.0 & 93.7 & 97.6 & \underline{58.3} & 56.5 & 72.8 \\
          & MambaAD~\cite{he2024mambaad}\dag  & 92.9 & 96.2 & 93.0 & 97.6 & 51.2 & 55.1 & 77.3  \\
          & \textbf{Dinomaly} (Ours) &\textbf{95.4} & \textbf{98.4} & \textbf{95.6} & \underline{97.8} & \textbf{70.1} & \textbf{68.0} & 76.5 \\
\hline
    \multirow{8}[0]{*}{Uni-Medical~\cite{zhang2024ader}} & RD4AD~\cite{deng2022anomaly} & 76.1  & 75.3  & 78.2  & 96.5  & 38.3  & 39.8  & \underline{86.8} \\
          & SimpleNet~\cite{liu2023simplenet} & 77.5  & 77.7  & 76.7  & 94.3  & 34.4  & 36.0  & 77.0\\
          & DeSTSeg~\cite{zhang2023destseg} & 78.5  & 77.0  & 78.2  & 65.7  & 41.7  & 34.0  & 35.3\\
          & UniAD~\cite{you2022unified}\dag & 79.0  & 76.1  & 77.1  & 96.6 & 39.3  & 41.1  & 86.0 \\
          & DiAD~\cite{he2024diffusion}\dag  & 78.8  & 77.2  & 77.7  & 95.8  & 34.2  & 35.5  & 84.3\\
          & ViTAD~\cite{zhang2023exploring}\dag & 81.8 & 80.7 & 80.0 & \textbf{97.1} & \underline{48.3} & \underline{48.2} & 86.7 \\
          & MambaAD~\cite{he2024mambaad}\dag  & \underline{83.9} & \underline{80.8} & \textbf{81.9} & \underline{96.8} & 45.8 & 47.5 & \textbf{88.2}  \\
          & \textbf{Dinomaly} (Ours) &\textbf{84.9} & \textbf{84.1} & \underline{81.0} & \underline{96.8} & \textbf{51.7} & \textbf{52.1} & 85.5 \\
          \bottomrule
\end{tabular}
}
\label{tab:mpdd}
\end{table*}

 \textbf{Ablations on VisA.} Similar to Table~\ref{tab3} that conduct ablation experiments on MVTec-AD, we additionally run them on VisA for further validations. As shown in Table~\ref{tab:abvisa}, proposed components of Dinomaly contribute to the AD performances on VisA as on MVTec-AD.
 
\textbf{Noisy Rates}. We conduct ablations on the discarding rate of the Dropouts in MLP bottleneck, as shown in Table~\ref{tab:noise}. Experimental results demonstrate that Dinomaly is robust to different levels of dropout rate.
 
 \textbf{Reconstruction Constraint.} We quantitatively examine different reconstruction schemes presented in Figure~\ref{fig4}. As shown in Table~\ref{tab:scheme}, group-to-group LC outperforms layer-to-layer supervision.  On image-level metrics, 1-group LC with all layers added performs similarly to its 2-group counterpart that separates low-level and high-level layers; however, 1-group LC mixes low-level and high-level features which is harmful for anomaly localization. More ablations on scalability, input size, pre-trained foundations, etc., are presented in Appendix~\ref{sec:ablation}.

\textbf{Attention \textit{vs.} Convolution.} Previous works and this paper have proposed to leverage attentions instead of convolutions in UAD. Here, we conduct experiments substituting the attention in the decoder of Dinomaly by convolutions as the spatial mixers. Following MetaFormer \cite{yu2023metaformer}, we employ Inverted Bottleneck block that consists of $1\times 1$ conv, GELU activation, $N\times N$ deep-wise conv, and $1\times 1$ conv, sequentially. The results are shown in Table~\ref{tab:conv}, where Attentions outperform Convolutions, especially for pixel-level anomaly localization. In addition, utilizing convolutions in the decoder can still yield SoTA results, demonstrating the universality of the proposed Dinomaly.

\textbf{Neighbour-Masking.} Prior method \cite{you2022unified} proposed to mask the keys and values in an $n\times n$ square centered at each query, in order to alleviate identity mapping in Attention. This mechanism can also be applied to Linear Attention as well. As shown in Table~\ref{tab:conv}, neighbor-masking can further improve Dinomaly with both Softmax Attention and Linear Attention moderately.

\textbf{Noise Bottleneck.} UniAD \cite{you2022unified} proposed to perturb the encoder features by Feature Jitter, i.e. adding Gaussian noise with \textit{scale} to control the noise magnitude. It is also easy to borrow the masking strategy of MAE \cite{he2022masked} to randomly mask patch tokens before the decoder. We evaluate the effectiveness of feature jitter and patch-masking in Dinomaly by placing it at the beginning of Noisy Bottleneck. As shown in Table~\ref{tab:dpfj}, both Dropout and Feature Jitter can be a good noise injector in Noisy Bottleneck. Meanwhile, Dropout is more robust to the noisy scale hyperparameter, and more elegant without introducing new modules.

\textbf{Adaptation on CNN Method.} Some proposed elements (Linear Attention and Loose Constraint) are closely bounded to modern ViTs. Loose Loss (hard-mining) can be directly applied to previous CNN-based methods, e.g., RD4AD \cite{deng2022anomaly}. Noisy Bottleneck can be adapted to RD4AD with minor modifications (apply dropout before MFF layer). We apply these modules to RD4AD to validate the effectiveness of our contributions. The results are shown in Table~\ref{tab:cnn}, where these two elements boost the performance of RD4AD to a whole new level that can be compared with prior MUAD SoTAs.

\textbf{Scaling of Compared Method.} As previously discussed in the Experiment section, compared method cannot fully utilize the scaling of pre-trained method, model size, and input size. For example, RD4AD \cite{deng2022anomaly} found WideResNet50 better than WideResNet101 as the encoder backbone. ViTAD \cite{zhang2023exploring} found ViT-Small better than ViT-Base. Here, we also present the experiments on pre-training method and input size of ViTAD, as shown in Table~\ref{tab:vitad}. It is also noted that the paradigm of ViTAD is very similar to RD4AD (replacing CNN by ViT) as well as the starting point of Dinomaly (the first row in the ablation Table~\ref{tab3}).

\textbf{Computation Comparison.} The computation costs of Dinomaly variants were previously presented in Table~\ref{tab:arch} and Table~\ref{tab:resolution}. Here, we compare the computation consumption of Dinomaly and prior works. As shown in Table~\ref{tab:cost}, Dinomaly can be easily scaled by model size and input size to match different application scenarios.

\textbf{Random Seeds.} Due to limited computation resources, experiments in this paper are conducted for one run with random seed=1. Here, we conduct 5 runs with 5 random seeds on MVTec-AD. As shown in Table~\ref{tab:seed}, Dinomaly is robust to randomness.

\section{Additional Dataset}
To demonstrate the generalization of our method, we conduct experiments on three more popular anomaly detection datasets under MUAD setting, including MPDD and BTAD and Uni-Medical. The MPDD \cite{jezek2021deep} (Metal Parts Defect Detection Dataset ) is a dataset aimed at benchmarking visual defect detection methods in industrial metal parts manufacturing. It consists of more than 1346 images across 6 categories with pixel-precise defect annotation masks. The BTAD \cite{mishra2021vt} ( beanTech Anomaly Detection) dataset is a real-world industrial anomaly dataset. The dataset contains a total of 2830 real-world images of 3 industrial products showcasing body and surface defects. It is noted that the training set of BTAD is noisy because it contains anomalous samples \cite{jiang2022softpatch}. Uni-Medical \cite{zhang2024ader} is a medical UAD dataset consisting of 2D image slices from 3D CT volumes. It contains 13339 training images and 7013 test images across three objects, i.e., brain CT, liver CT, and retinal OCT. This dataset is not entirely suitable for evaluating 2D anomaly detection methods, as identifying lesions in medical images requires 3D contextual information. The training hyperparameters are the same to MVTec-AD, except the dropout rate for Uni-Medical is increased to 0.4.  
The performance of Dinomaly and previous SoTAs is presented in Table~\ref{tab:mpdd}, where our method demonstrates superior results.

\section{Results Per-Category}
\label{sec:perclass}

For future research, we report the per-class results of MVTec-AD~\cite{bergmann2019mvtec}, VisA~\cite{zou2022spot}, and Real-IAD~\cite{wang2024real}. The performance of compared methods is drawn from MambaAD \cite{he2024mambaad}. Thanks for their exhaustive reproducing. The results of image-level anomaly detection and pixel-level anomaly localization on MVTec-AD are presented in Table~\ref{tab:mvtecsp} and Table~\ref{tab:mvtecpx}, respectively. The results of image-level anomaly detection and pixel-level anomaly localization on VisA are presented in Table~\ref{tab:visasp} and Table~\ref{tab:visapx}, respectively. The results of image-level anomaly detection and pixel-level anomaly localization on Real-IAD are presented in Table~\ref{tab:realiadsp} and Table~\ref{tab:realiadpx}, respectively.

\section{Qualitative Visualization}
\label{sec:visual}

\begin{figure*}[!h]
\centerline{\includegraphics[width=\textwidth]{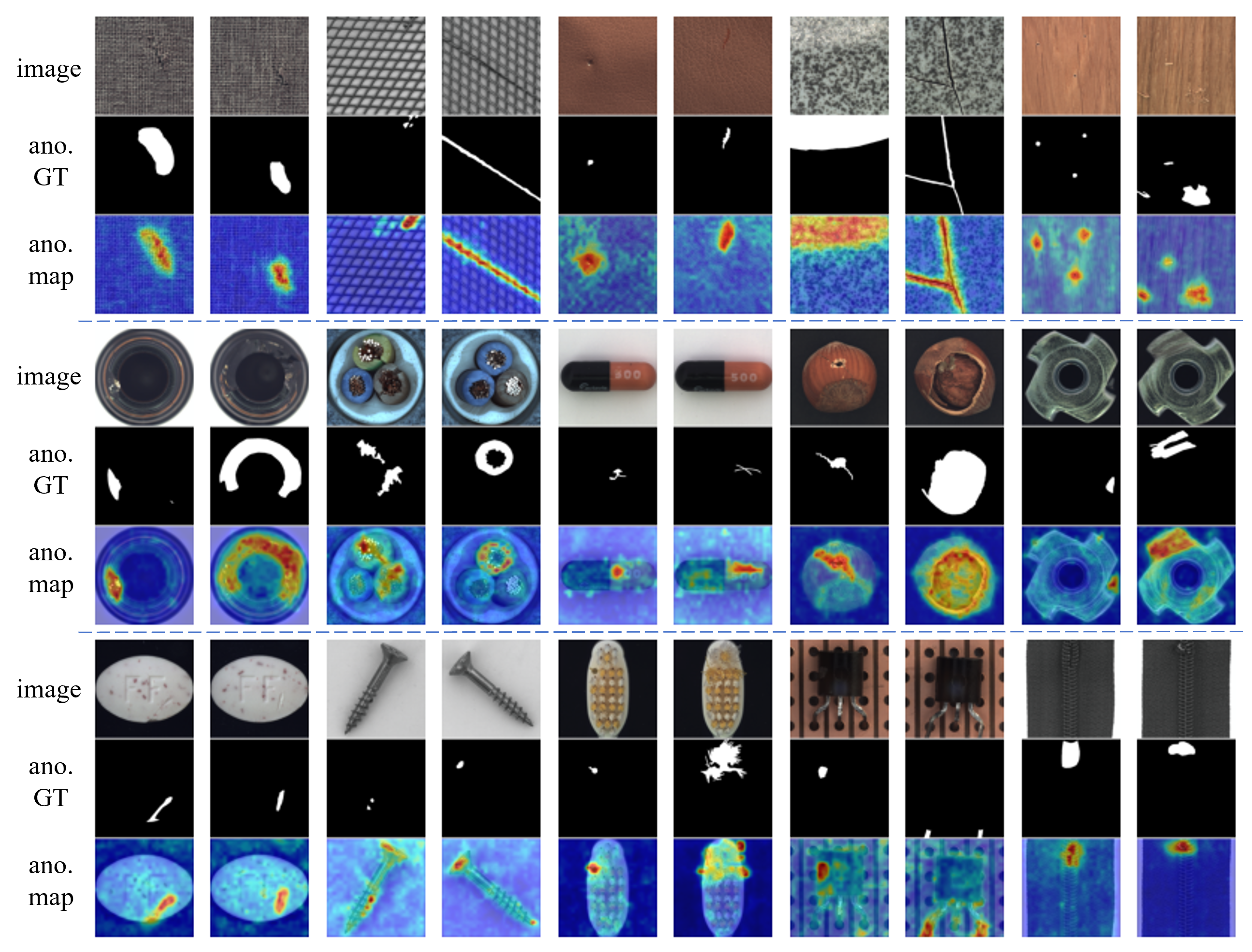}}
\caption{Anomaly maps visualization on MVTec-AD. All samples are randomly chosen.}
\label{fig_mvtec}
\end{figure*}

\begin{figure*}[!h]
\centerline{\includegraphics[width=\textwidth]{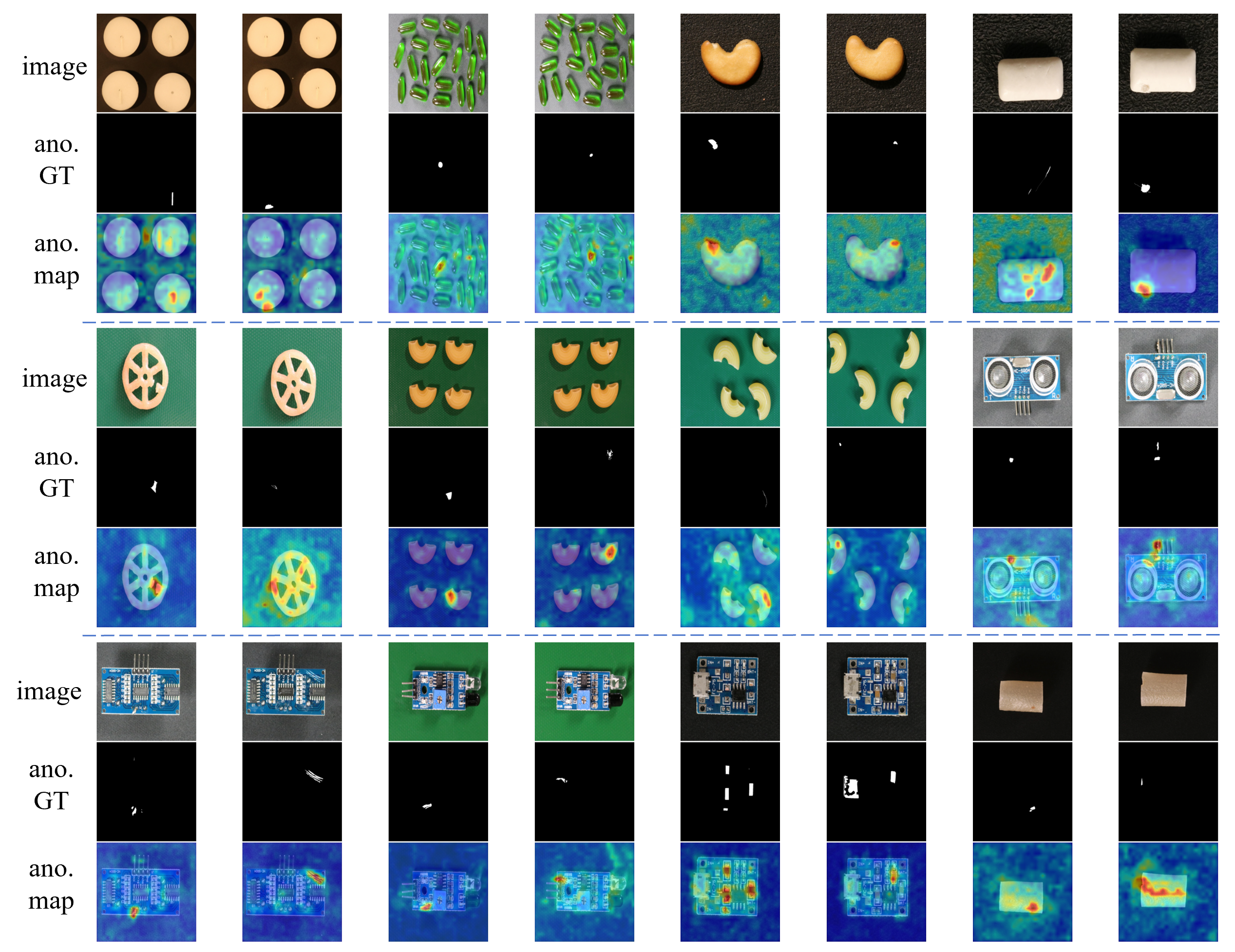}}
\caption{Anomaly maps visualization on VisA. All samples are randomly chosen.}
\label{fig_visa}
\end{figure*}

\begin{figure*}[!h]
\centerline{\includegraphics[width=\textwidth]{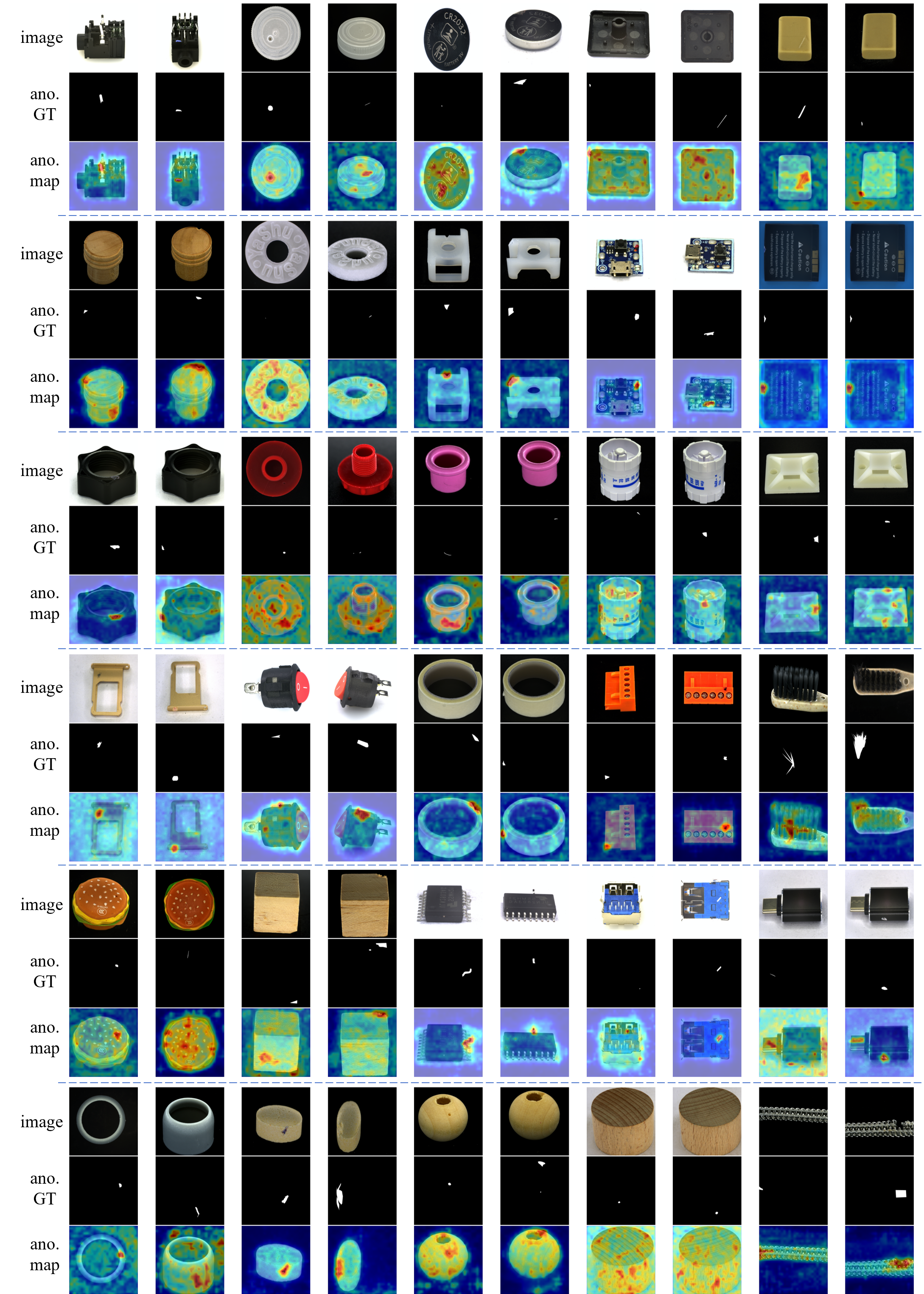}}
\caption{Anomaly maps visualization on Real-IAD. All samples are randomly chosen.}
\label{fig_realiad}
\end{figure*}

We visualize the output anomaly maps of Dinomaly on MVTec-AD, VisA, and Real-IAD, as shown in Figure~\ref{fig_mvtec}, Figure~\ref{fig_visa}, and Figure~\ref{fig_realiad}. It is noted that all visualized samples are randomly chosen without artificial selection.  

\section{Limitation}
Vision Transformers are known for their high computation cost, which can be a barrier to low-computation scenarios that require inference speed. Future research can be conducted on the efficiency of Transformer-based methods, such as distillation, pruning, and hardware-friendly attention mechanism (such as FlashAttention).

As discussed in section \ref{sec:relatedwork}, Dinomaly is used for sensory AD that aims to detect regional anomalies in normal backgrounds. It is not suitable for semantic AD. Previous works have shown that methods designed for sensory AD usually fail to be competitive under semantic AD tasks \cite{you2022unified,deng2022anomaly}. Conversely, methods designed for semantic AD do not perform well on sensory AD tasks \cite{reiss2021panda,reiss2022anomaly}. Future work can be conducted to unify these two tasks, but according to the "no free lunch" theorem, we believe that methods designed for specific anomaly assumption are likely to be more convincing. 

Other special UAD settings, such as zero-shot UAD (vision-language model based) \cite{jeong2023winclip}, few-shot UAD \cite{huang2022registration}, UAD under noisy training set \cite{jiang2022softpatch}, are not included in this work. 

\begin{table*}[!h]
  \centering
  \caption{Per-class performance on\textbf{ MVTec-AD} dataset for multi-class anomaly detection with AUROC/AP/$F_1$-max metrics.}
  \resizebox{1\linewidth}{!}{
    \begin{tabular}{p{3em}<{\centering} p{3.25em}<{\centering}p{6.2em}<{\centering} p{6.2em}<{\centering} p{6.2em}<{\centering} p{6.2em}<{\centering} p{6.2em}<{\centering} p{6.2em}<{\centering} p{6.2em}<{\centering} p{6.2em}<{\centering} p{7em}<{\centering}}
    \toprule
    \multicolumn{2}{c}{Method~$\rightarrow$} & RD4AD~\cite{deng2022anomaly} & UniAD~\cite{you2022unified} & SimpleNet~\cite{liu2023simplenet}  & DeSTSeg~\cite{zhang2023destseg}  & DiAD~\cite{he2024diffusion} & MambaAD~\cite{he2024mambaad} & \cellcolor{tab_ours}Dinomaly \\
    \cline{1-2}
    \multicolumn{2}{c}{Category~$\downarrow$} & CVPR'22 & NeurlPS'22 & CVPR'23 & CVPR'23 & AAAI'24& Arxiv'24 & \cellcolor{tab_ours}Ours \\
    \hline
    \multicolumn{1}{c}{\multirow{10}[1]{*}{\begin{turn}{-90}Objects\end{turn}}} & \multicolumn{1}{c}{Bottle} & 99.6/99.9/98.4 & 99.7/\textbf{100.}/\textbf{100.} & \textbf{100.}/\textbf{100.}/\textbf{100.} & 98.7/99.6/96.8  & 99.7/96.5/91.8 &\textbf{100.}/\textbf{100.}/\textbf{100.} & \cellcolor{tab_ours}\textbf{100.}/\textbf{100.}/\textbf{100.} \\
    
    \multicolumn{1}{c}{} & \multicolumn{1}{c}{\cellcolor{tab_others}Cable} & \cellcolor{tab_others}84.1/89.5/82.5 & \cellcolor{tab_others}95.2/95.9/88.0 & \cellcolor{tab_others}97.5/98.5/94.7 & \cellcolor{tab_others}89.5/94.6/85.9  & \cellcolor{tab_others}94.8/98.8/95.2 &\cellcolor{tab_others}98.8/99.2/95.7 & \cellcolor{tab_ours}\textbf{100.}/\textbf{100.}/\textbf{100.} \\
    
    \multicolumn{1}{c}{} & \multicolumn{1}{c}{Capsule}  & 94.1/96.9/96.9 & 86.9/97.8/94.4 & 90.7/97.9/93.5 & 82.8/95.9/92.6  & 89.0/97.5/95.5 &94.4/98.7/94.9 & \cellcolor{tab_ours}\textbf{97.9}/\textbf{99.5}/\textbf{97.7}\\
    
    \multicolumn{1}{c}{} & \multicolumn{1}{c}{\cellcolor{tab_others}Hazelnut} & \cellcolor{tab_others}60.8/69.8/86.4 & \cellcolor{tab_others}99.8/\textbf{100.}/99.3 & \cellcolor{tab_others}99.9/99.9/99.3 & \cellcolor{tab_others}98.8/99.2/98.6  & \cellcolor{tab_others}99.5/99.7/97.3 &\cellcolor{tab_others}\textbf{100.}/\textbf{100.}/\textbf{100.} & \cellcolor{tab_ours}\textbf{100.}/\textbf{100.}/\textbf{100.}\\
    
    \multicolumn{1}{c}{} & \multicolumn{1}{c}{Metal Nut}  & \textbf{100.}/\textbf{100.}/99.5 & 99.2/99.9/99.5 & 96.9/99.3/96.1 & 92.9/98.4/92.2  & 99.1/96.0/91.6 &99.9/\textbf{100.}/99.5 & \cellcolor{tab_ours}\textbf{100.}/\textbf{100.}/\textbf{100.} \\
    
    \multicolumn{1}{c}{} & \multicolumn{1}{c}{\cellcolor{tab_others}Pill}  & \cellcolor{tab_others}97.5/99.6/96.8 & \cellcolor{tab_others}93.7/98.7/95.7 & \cellcolor{tab_others}88.2/97.7/92.5 & \cellcolor{tab_others}77.1/94.4/91.7  & \cellcolor{tab_others}95.7/98.5/94.5 &\cellcolor{tab_others}97.0/99.5/96.2 & \cellcolor{tab_ours}\textbf{99.1}/\textbf{99.9}/\textbf{98.3}\\
    
    \multicolumn{1}{c}{} & \multicolumn{1}{c}{Screw} & 97.7/99.3/95.8 & 87.5/96.5/89.0 & 76.7/90.6/87.7 & 69.9/88.4/85.4  & 90.7/\textbf{99.7}/\textbf{97.9} & 94.7/97.9/94.0 & \cellcolor{tab_ours}\textbf{98.4}/99.5/96.1\\
    
    \multicolumn{1}{c}{} & \multicolumn{1}{c}{\cellcolor{tab_others}Toothbrush} & \cellcolor{tab_others}97.2/99.0/94.7 & \cellcolor{tab_others}94.2/97.4/95.2& \cellcolor{tab_others}89.7/95.7/92.3 & \cellcolor{tab_others}71.7/89.3/84.5  & \cellcolor{tab_others}99.7/99.9/99.2&\cellcolor{tab_others}98.3/99.3/98.4 & \cellcolor{tab_ours}\textbf{100.}/\textbf{100.}/\textbf{100.} \\
    
    \multicolumn{1}{c}{} & \multicolumn{1}{c}{Transistor} & 94.2/95.2/90.0 & 99.8/98.0/93.8 & 99.2/98.7/97.6 & 78.2/79.5/68.8  & 99.8/99.6/97.4 &\textbf{100.}/\textbf{100.}/\textbf{100.} & \cellcolor{tab_ours}99.0/98.0/96.4\\
    
    \multicolumn{1}{c}{} & \multicolumn{1}{c}{\cellcolor{tab_others}Zipper}  & \cellcolor{tab_others}99.5/99.9/99.2 & \cellcolor{tab_others}95.8/99.5/97.1 & \cellcolor{tab_others}99.0/99.7/98.3 & \cellcolor{tab_others}88.4/96.3/93.1  & \cellcolor{tab_others}95.1/99.1/94.4 &\cellcolor{tab_others}99.3/99.8/97.5 & \cellcolor{tab_ours}\textbf{100.}/\textbf{100.}/\textbf{100.}\\
    \hline
    \multicolumn{1}{c}{\multirow{5}[1]{*}{\begin{turn}{-90}Textures\end{turn}}} & \multicolumn{1}{c}{Carpet}  & 98.5/99.6/97.2 & \textbf{99.8}/99.9/\textbf{99.4} & 95.7/98.7/93.2 & 95.9/98.8/94.9  & 99.4/99.9/98.3 &\textbf{99.8}/99.9/\textbf{99.4}  & \cellcolor{tab_ours}\textbf{99.8}/\textbf{100.}/98.9\\
    
    \multicolumn{1}{c}{} & \multicolumn{1}{c}{\cellcolor{tab_others}Grid} & \cellcolor{tab_others}98.0/99.4/96.5 & \cellcolor{tab_others}98.2/99.5/97.3 & \cellcolor{tab_others}97.6/99.2/96.4 & \cellcolor{tab_others}97.9/99.2/96.6  & \cellcolor{tab_others}98.5/99.8/97.7 &\textbf{100.}/\cellcolor{tab_others}\textbf{100.}/\textbf{100.} & \cellcolor{tab_ours}99.9/\textbf{100.}/99.1\\
    
    \multicolumn{1}{c}{} & \multicolumn{1}{c}{Leather} & \textbf{100.}/\textbf{100.}/\textbf{100.} & \textbf{100.}/\textbf{100.}/\textbf{100.} & \textbf{100.}/\textbf{100.}/\textbf{100.} & 99.2/99.8/98.9  & 99.8/99.7/97.6 &\textbf{100.}/\textbf{100.}/\textbf{100.} & \cellcolor{tab_ours}\textbf{100.}/\textbf{100.}/\textbf{100.}\\
    
    \multicolumn{1}{c}{} & \multicolumn{1}{c}{\cellcolor{tab_others}Tile}  & \cellcolor{tab_others}98.3/99.3/96.4 & \cellcolor{tab_others}99.3/99.8/98.2 & \cellcolor{tab_others}99.3/99.8/98.8 & \cellcolor{tab_others}97.0/98.9/95.3  & \cellcolor{tab_others}96.8/99.9/98.4 &\cellcolor{tab_others}98.2/99.3/95.4 & \cellcolor{tab_ours}\textbf{100.}/\textbf{100.}/\textbf{100.}\\
    
    \multicolumn{1}{c}{} & \multicolumn{1}{c}{Wood}  & 99.2/99.8/98.3 & 98.6/99.6/96.6 & 98.4/99.5/96.7 & 99.9/\textbf{100.}/99.2  & 99.7/\textbf{100.}/\textbf{100.} &98.8/99.6/96.6 & \cellcolor{tab_ours}99.8/99.9/99.2\\
    \hline
    \multicolumn{2}{c}{\cellcolor{tab_others}Mean} & \cellcolor{tab_others}94.6/96.5/95.2 & \cellcolor{tab_others}96.5/98.8/96.2 & \cellcolor{tab_others}95.3/98.4/95.8 & \cellcolor{tab_others}89.2/95.5/91.6  & \cellcolor{tab_others}97.2/99.0/96.5 &\cellcolor{tab_others}98.6/99.6/97.8  &  \cellcolor{tab_ours}\textbf{99.6}/\textbf{99.8}/\textbf{99.0}\\
    \bottomrule
    \end{tabular}%
  }
  \label{tab:mvtecsp}%
\end{table*}%

\begin{table*}[!h]
\centering
\caption{Per-class performance on \textbf{MVTec-AD} dataset for multi-class anomaly localization with AUROC/AP/$F_1$-max/AUPRO metrics.}
\resizebox{\linewidth}{!}{
\begin{tabular}{ccccccccc}
\toprule
\multicolumn{2}{c}{Method~$\rightarrow$} & RD4AD~\cite{deng2022anomaly} & UniAD~\cite{you2022unified} & SimpleNet~\cite{liu2023simplenet} & DeSTSeg~\cite{zhang2023destseg} & DiAD~\cite{he2024diffusion} & MambaAD~\cite{he2024mambaad} & \cellcolor{tab_ours}Dinomaly \\
\cline{1-2}
\multicolumn{2}{c}{Category~$\downarrow$} & CVPR'22 & NeurlPS'22 & CVPR'23 & CVPR'23 & AAAI'24& Arxiv'24 & \cellcolor{tab_ours}Ours \\
\hline
\multicolumn{1}{c}{\multirow{10}[1]{*}{\begin{turn}{-90}Objects\end{turn}}} & \multicolumn{1}{c}{Bottle} & 97.8/68.2/67.6/94.0 & 98.1/66.0/69.2/93.1 & 97.2/53.8/62.4/89.0 & 93.3/61.7/56.0/67.5 & 98.4/52.2/54.8/86.6 & 98.8/79.7/76.7/95.2 & \cellcolor{tab_ours}\textbf{99.2}/\textbf{88.6}/\textbf{84.2}/\textbf{96.6} \\

\multicolumn{1}{c}{} & \multicolumn{1}{c}{\cellcolor{tab_others}Cable} & \cellcolor{tab_others}85.1/26.3/33.6/75.1 & \cellcolor{tab_others}97.3/39.9/45.2/86.1 & \cellcolor{tab_others}96.7/42.4/51.2/85.4 & \cellcolor{tab_others}89.3/37.5/40.5/49.4 & \cellcolor{tab_others}96.8/50.1/57.8/80.5 & \cellcolor{tab_others}95.8/42.2/48.1/90.3 & \cellcolor{tab_ours}\textbf{98.6}/\textbf{72.0}/\textbf{74.3}/\textbf{94.2} \\

\multicolumn{1}{c}{} & \multicolumn{1}{c}{Capsule} & \textbf{98.8}/43.4/50.0/94.8 & 98.5/42.7/46.5/92.1 &98.5/35.4/44.3/84.5 & 95.8/47.9/48.9/62.1 & 97.1/42.0/45.3/87.2 & 98.4/43.9/47.7/92.6 & 98.7/\cellcolor{tab_ours}\textbf{61.4}/\textbf{60.3}/\textbf{97.2} \\

\multicolumn{1}{c}{} & \multicolumn{1}{c}{\cellcolor{tab_others}Hazelnut} & \cellcolor{tab_others}97.9/36.2/51.6/92.7 & \cellcolor{tab_others}98.1/55.2/56.8/94.1 &\cellcolor{tab_others}98.4/44.6/51.4/87.4 & \cellcolor{tab_others}98.2/65.8/61.6/84.5 & \cellcolor{tab_others}98.3/79.2/\textbf{80.4}/91.5& \cellcolor{tab_others}99.0/63.6/64.4/95.7 & \cellcolor{tab_ours}\textbf{99.4}/\textbf{82.2}/76.4/\textbf{97.0} \\

\multicolumn{1}{c}{} & \multicolumn{1}{c}{Metal Nut} & 94.8/55.5/66.4/91.9 & 62.7/14.6/29.2/81.8 &\textbf{98.0}/\textbf{83.1}/79.4/85.2 & 84.2/42.0/22.8/53.0 & 97.3/30.0/38.3/90.6 & 96.7/74.5/79.1/93.7 & \cellcolor{tab_ours}96.9/78.6/\textbf{86.7}/\textbf{94.9} \\

\multicolumn{1}{c}{} & \multicolumn{1}{c}{\cellcolor{tab_others}Pill} & \cellcolor{tab_others}97.5/63.4/65.2/95.8 & \cellcolor{tab_others}95.0/44.0/53.9/95.3 &\cellcolor{tab_others}96.5/72.4/67.7/81.9 & \cellcolor{tab_others}96.2/61.7/41.8/27.9 & \cellcolor{tab_others}95.7/46.0/51.4/89.0 & \cellcolor{tab_others}97.4/64.0/66.5/95.7 & \cellcolor{tab_ours}\textbf{97.8}/\textbf{76.4}/\textbf{71.6}/\textbf{97.3} \\

\multicolumn{1}{c}{} & \multicolumn{1}{c}{Screw} & 99.4/40.2/44.6/96.8 & 98.3/28.7/37.6/95.2 & 96.5/15.9/23.2/84.0 & 93.8/19.9/25.3/47.3 & 97.9/\textbf{60.6}/\textbf{59.6}/95.0 & 99.5/49.8/50.9/97.1 & \cellcolor{tab_ours}\textbf{99.6}/60.2/\textbf{59.6}/\textbf{98.3} \\

\multicolumn{1}{c}{} & \multicolumn{1}{c}{\cellcolor{tab_others}Toothbrush} & \cellcolor{tab_others}\textbf{99.0}/53.6/58.8/92.0 & \cellcolor{tab_others}98.4/34.9/45.7/87.9& \cellcolor{tab_others}98.4/46.9/52.5/87.4 & \cellcolor{tab_others}96.2/52.9/58.8/30.9 & \cellcolor{tab_others}\textbf{99.0}/78.7/72.8/95.0 & \cellcolor{tab_others}\textbf{99.0}/48.5/59.2/91.7 & \cellcolor{tab_ours} 98.9/\textbf{51.5}/\textbf{62.6}/\textbf{95.3} \\

\multicolumn{1}{c}{} & \multicolumn{1}{c}{Transistor} & 85.9/42.3/45.2/74.7 & \textbf{97.9}/59.5/\textbf{64.6}/\textbf{93.5} & 95.8/58.2/56.0/83.2 & 73.6/38.4/39.2/43.9 & 95.1/15.6/31.7/90.0 &96.5/69.4/67.1/87.0 & \cellcolor{tab_ours}93.2/\textbf{59.9}/58.5/77.0 \\

\multicolumn{1}{c}{} & \multicolumn{1}{c}{\cellcolor{tab_others}Zipper} & \cellcolor{tab_others}98.5/53.9/60.3/94.1 &\cellcolor{tab_others} 96.8/40.1/49.9/92.6 & \cellcolor{tab_others}97.9/53.4/54.6/90.7 & \cellcolor{tab_others}97.3/64.7/59.2/66.9 & \cellcolor{tab_others}96.2/60.7/60.0/91.6 & \cellcolor{tab_others}98.4/60.4/61.7/94.3 & \cellcolor{tab_ours}\textbf{99.2}/\textbf{79.5}/\textbf{75.4}/\textbf{97.2} \\
\hline

\multicolumn{1}{c}{\multirow{5}[1]{*}{\begin{turn}{-90}Textures\end{turn}}} & \multicolumn{1}{c}{Carpet} & 99.0/58.5/60.4/95.1 & 98.5/49.9/51.1/94.4 & 97.4/38.7/43.2/90.6 & 93.6/59.9/58.9/89.3 & 98.6/42.2/46.4/90.6 &99.2/60.0/63.3/96.7 & \cellcolor{tab_ours}\textbf{99.3}/\textbf{68.7}/\textbf{71.1}/\textbf{97.6} \\

\multicolumn{1}{c}{} & \multicolumn{1}{c}{\cellcolor{tab_others}Grid} & \cellcolor{tab_others}96.5/23.0/28.4/97.0 & \cellcolor{tab_others}63.1/10.7/11.9/92.9 & \cellcolor{tab_others}96.8/20.5/27.6/88.6/ & \cellcolor{tab_others}97.0/42.1/46.9/86.8 & \cellcolor{tab_others}96.6/66.0/64.1/94.0 & \cellcolor{tab_others}99.2/47.4/47.7/97.0 & \cellcolor{tab_ours}\textbf{99.4}/\textbf{55.3}/\textbf{57.7}/\textbf{97.2}\\

\multicolumn{1}{c}{} & \multicolumn{1}{c}{Leather} & 99.3/38.0/45.1/97.4 & 98.8/32.9/34.4/96.8 &98.7/28.5/32.9/92.7 & \textbf{99.5}/\textbf{71.5}/\textbf{66.5}/91.1 & 98.8/56.1/62.3/91.3 & 99.4/50.3/53.3/98.7 & \cellcolor{tab_ours}99.4/52.2/55.0/\textbf{97.6} \\

\multicolumn{1}{c}{} & \multicolumn{1}{c}{\cellcolor{tab_others}Tile} & \cellcolor{tab_others}95.3/48.5/60.5/85.8 & \cellcolor{tab_others}91.8/42.1/50.6/78.4 & \cellcolor{tab_others}95.7/60.5/59.9/90.6 & \cellcolor{tab_others}93.0/71.0/66.2/87.1 & \cellcolor{tab_others}92.4/65.7/64.1/\textbf{90.7} & \cellcolor{tab_others}93.8/45.1/54.8/80.0 & \cellcolor{tab_ours}\textbf{98.1}/\textbf{80.1}/\textbf{75.7}/90.5 \\

\multicolumn{1}{c}{} & \multicolumn{1}{c}{Wood} & 95.3/47.8/51.0/90.0 & 93.2/37.2/41.5/86.7 &91.4/34.8/39.7/76.3 & 95.9/\textbf{77.3}/\textbf{71.3}/83.4 & 93.3/43.3/43.5/\textbf{97.5} & 94.4/46.2/48.2/91.2 & \cellcolor{tab_ours}\textbf{97.6}/72.8/68.4/94.0 \\
\hline

\multicolumn{2}{c}{\cellcolor{tab_others}Mean} & \cellcolor{tab_others}96.1/48.6/53.8/91.1 & \cellcolor{tab_others}96.8/43.4/49.5/90.7 & \cellcolor{tab_others}96.9/45.9/49.7/86.5 & \cellcolor{tab_others}93.1/54.3/50.9/64.8 & \cellcolor{tab_others}96.8/52.6/55.5/90.7 & \cellcolor{tab_others}97.7/56.3/59.2/93.1 & \cellcolor{tab_ours}\textbf{98.4}/\textbf{69.3}/\textbf{69.2}/\textbf{94.8} \\
\bottomrule
\end{tabular}
}
\label{tab:mvtecpx}%
\end{table*}%

\begin{table*}[!h]
\centering
\caption{Per-class performance on \textbf{VisA} dataset for multi-class anomaly detection with AUROC/AP/F1-max metrics.}
\resizebox{1.\linewidth}{!}{
\begin{tabular}{cccccccc}
\toprule
Method~$\rightarrow$ & RD4AD~\cite{deng2022anomaly} & UniAD~\cite{you2022unified} & SimpleNet~\cite{liu2023simplenet} & DeSTSeg~\cite{zhang2023destseg} & DiAD~\cite{he2024diffusion} & MambaAD & \cellcolor{tab_ours}Dinomaly \\
\cline{1-1}
Category~$\downarrow$ & CVPR'22 & NeurlPS'22 & CVPR'23 & CVPR'23 & AAAI'24& Arxiv'24 & \cellcolor{tab_ours}Ours \\
\hline

pcb1 & 96.2/95.5/91.9 & 92.8/92.7/87.8 & 91.6/91.9/86.0 & 87.6/83.1/83.7 & 88.1/88.7/80.7 & 95.4/93.0/91.6 & \cellcolor{tab_ours}\textbf{99.1}/\textbf{99.1}/\textbf{96.6} \\

\cellcolor{tab_others}pcb2 & \cellcolor{tab_others}97.8/97.8/94.2 & \cellcolor{tab_others}87.8/87.7/83.1 &\cellcolor{tab_others}92.4/93.3/84.5 &\cellcolor{tab_others}86.5/85.8/82.6 & \cellcolor{tab_others}91.4/91.4/84.7 &\cellcolor{tab_others}94.2/93.7/89.3 & \cellcolor{tab_ours}\textbf{99.3}/\textbf{99.2}/\textbf{97.0} \\

pcb3 & 96.4/96.2/91.0 & 78.6/78.6/76.1 & 89.1/91.1/82.6 & 93.7/95.1/87.0 & 86.2/87.6/77.6 & 93.7/94.1/86.7 & \cellcolor{tab_ours}\textbf{98.9}/\textbf{98.9}/\textbf{96.1} \\

\cellcolor{tab_others}pcb4 & \cellcolor{tab_others}99.9/99.9/99.0 & \cellcolor{tab_others}98.8/98.8/94.3 & \cellcolor{tab_others}97.0/97.0/93.5 &\cellcolor{tab_others}97.8/97.8/92.7 &\cellcolor{tab_others}99.6/99.5/97.0 &\cellcolor{tab_others}\textbf{99.9}/\textbf{99.9}/\textbf{98.5} & \cellcolor{tab_ours}99.8/99.8/98.0 \\
\hline

macaroni1 & 75.9/ 1.5/76.8 & 79.9/79.8/72.7 & 85.9/82.5/73.1 & 76.6/69.0/71.0 & 85.7/85.2/78.8 & 91.6/89.8/81.6 & \cellcolor{tab_ours}\textbf{98.0}/\textbf{97.6}/\textbf{94.2} \\

\cellcolor{tab_others}macaroni2 & \cellcolor{tab_others}88.3/84.5/83.8 & \cellcolor{tab_others}71.6/71.6/69.9 &\cellcolor{tab_others}68.3/54.3/59.7 & \cellcolor{tab_others}68.9/62.1/67.7 & \cellcolor{tab_others}62.5/57.4/69.6 &\cellcolor{tab_others}81.6/78.0/73.8 & \cellcolor{tab_ours}\textbf{95.9}/\textbf{95.7}/\textbf{90.7} \\

capsules & 82.2/90.4/81.3 & 55.6/55.6/76.9 &74.1/82.8/74.6 & 87.1/93.0/84.2 & 58.2/69.0/78.5 & 91.8/95.0/88.8 & \cellcolor{tab_ours}\textbf{98.6}/\textbf{99.0}/\textbf{97.1} \\

\cellcolor{tab_others}candle & \cellcolor{tab_others}92.3/92.9/86.0 & \cellcolor{tab_others}94.1/94.0/86.1 & \cellcolor{tab_others}84.1/73.3/76.6 & \cellcolor{tab_others}94.9/94.8/89.2 & \cellcolor{tab_others}92.8/92.0/87.6 &\cellcolor{tab_others}96.8/96.9/90.1 & \cellcolor{tab_ours}\textbf{98.7}/\textbf{98.8}/\textbf{95.1} \\
\hline

cashew & 92.0/95.8/90.7 & 92.8/92.8/91.4 & 88.0/91.3/84.7 & 92.0/96.1/88.1 & 91.5/95.7/89.7 & 94.5/97.3/91.1 & \cellcolor{tab_ours}\textbf{98.7}/\textbf{99.4}/\textbf{97.0} \\

\cellcolor{tab_others}chewinggum &\cellcolor{tab_others}94.9/97.5/92.1 &\cellcolor{tab_others}96.3/96.2/95.2 & \cellcolor{tab_others}96.4/98.2/93.8 & \cellcolor{tab_others}95.8/98.3/94.7 & \cellcolor{tab_others}99.1/99.5/95.9 &\cellcolor{tab_others}97.7/98.9/94.2 & \cellcolor{tab_ours}\textbf{99.8}/\textbf{99.9}/\textbf{99.0} \\

fryum & 95.3/97.9/91.5 & 83.0/83.0/85.0 & 88.4/93.0/83.3 & 92.1/96.1/89.5 & 89.8/95.0/87.2 & 95.2/97.7/90.5 & \cellcolor{tab_ours}\textbf{98.8}/\textbf{99.4}/\textbf{96.5} \\

\cellcolor{tab_others}pipe\_fryum &\cellcolor{tab_others}97.9/98.9/96.5 & \cellcolor{tab_others}94.7/94.7/93.9 &\cellcolor{tab_others}90.8/95.5/88.6 & \cellcolor{tab_others}94.1/97.1/91.9 & \cellcolor{tab_others}96.2/98.1/93.7 &\cellcolor{tab_others}98.7/99.3/97.0 & \cellcolor{tab_ours}\textbf{99.2}/\textbf{99.7}/\textbf{97.0} \\
\hline

Mean  & 92.4/92.4/89.6   & 85.5/85.5/84.4 & 87.2/87.0/81.8 & 88.9/89.0/85.2 & 86.8/88.3/85.1& 94.3/94.5/89.4 &   \cellcolor{tab_ours}\textbf{98.7}/\textbf{98.9}/\textbf{96.2} \\
\bottomrule
\end{tabular}
  }
  \label{tab:visasp}%
\end{table*}%

\begin{table*}[!h]
\centering
\caption{Per-class performance on \textbf{VisA} dataset for multi-class anomaly localization with AUROC/AP/$F_1$-max/AUPRO metrics.}
\resizebox{1\linewidth}{!}{
\begin{tabular}{cccccccc}
\toprule
Method~$\rightarrow$ & RD4AD~\cite{deng2022anomaly} & UniAD~\cite{you2022unified} & SimpleNet~\cite{liu2023simplenet} & DeSTSeg~\cite{zhang2023destseg} & DiAD~\cite{he2024diffusion} & MambaAD & \cellcolor{tab_ours}Dinomaly \\
\cline{1-1}
Category~$\downarrow$ & CVPR'22 & NeurlPS'22 & CVPR'23 & CVPR'23 & AAAI'24& Arxiv'24& \cellcolor{tab_ours} Ours \\
\midrule

pcb1 & 99.4/66.2/62.4/\textbf{95.8} & 93.3/ 3.9/ 8.3/64.1 & 99.2/86.1/78.8/83.6 & 95.8/46.4/49.0/83.2 & 98.7/49.6/52.8/80.2 &\textbf{99.8}/77.1/72.4/92.8 &\cellcolor{tab_ours}99.5/\textbf{87.9}/\textbf{80.5}/95.1 \\

\cellcolor{tab_others}pcb2 & \cellcolor{tab_others}98.0/22.3/30.0/90.8 & \cellcolor{tab_others}93.9/ 4.2/ 9.2/66.9 & \cellcolor{tab_others}96.6/ 8.9/18.6/85.7 & \cellcolor{tab_others}97.3/14.6/28.2/79.9 & \cellcolor{tab_others}95.2/ 7.5/16.7/67.0 &\cellcolor{tab_others}\textbf{98.9}/13.3/23.4/89.6 &\cellcolor{tab_ours}98.0/\textbf{47.0}/\textbf{49.8}/\textbf{91.3}\\

pcb3 & 97.9/26.2/35.2/93.9 & 97.3/13.8/21.9/70.6 & 97.2/31.0/36.1/85.1 & 97.7/28.1/33.4/62.4 & 96.7/ 8.0/18.8/68.9 &\textbf{99.1}/18.3/27.4/89.1 &\cellcolor{tab_ours}98.4/\textbf{41.7}/\textbf{45.3}/\textbf{94.6}\\

\cellcolor{tab_others}pcb4 &\cellcolor{tab_others}97.8/31.4/37.0/88.7 & \cellcolor{tab_others}94.9/14.7/22.9/72.3 &\cellcolor{tab_others}93.9/23.9/32.9/61.1 &\cellcolor{tab_others}95.8/\textbf{53.0}/\textbf{53.2}/76.9 & \cellcolor{tab_others}97.0/17.6/27.2/85.0 &\cellcolor{tab_others}98.6/47.0/46.9/87.6 &\cellcolor{tab_ours}\textbf{98.7}/50.5/53.1/\textbf{94.4}\\

\midrule
macaroni1 & 99.4/ 2.9/6.9/95.3 & 97.4/ 3.7/ 9.7/84.0 & 98.9/ 3.5/8.4/92.0 &99.1/ 5.8/13.4/62.4 & 94.1/10.2/16.7/68.5 &99.5/17.5/27.6/95.2 &\cellcolor{tab_ours}\textbf{99.6}/\textbf{33.5}/\textbf{40.6}/\textbf{96.4}\\

\cellcolor{tab_others}macaroni2 & \cellcolor{tab_others}99.7/13.2/21.8/97.4 & \cellcolor{tab_others}95.2/ 0.9/ 4.3/76.6 & \cellcolor{tab_others}93.2/ 0.6/ 3.9/77.8 & \cellcolor{tab_others}98.5/ 6.3/14.4/70.0 & \cellcolor{tab_others}93.6/ 0.9/ 2.8/73.1 &\cellcolor{tab_others}99.5/ 9.2/16.1/96.2 &\cellcolor{tab_ours}\textbf{99.7}/\textbf{24.7}/\textbf{36.1}/\textbf{98.7}\\

capsules &99.4/60.4/60.8/93.1 & 88.7/ 3.0/ 7.4/43.7 &97.1/52.9/53.3/73.7 &96.9/33.2/ 9.1/76.7 & 97.3/10.0/21.0/77.9 &99.1/61.3/59.8/91.8 &\cellcolor{tab_ours}\textbf{99.6}/\textbf{65.0}/\textbf{66.6}/\textbf{97.4}\\

\cellcolor{tab_others}candle & \cellcolor{tab_others}99.1/25.3/35.8/94.9 & \cellcolor{tab_others}98.5/17.6/27.9/91.6 & \cellcolor{tab_others}97.6/ 8.4/16.5/87.6 & \cellcolor{tab_others}98.7/39.9/45.8/69.0 & \cellcolor{tab_others}97.3/12.8/22.8/89.4 &\cellcolor{tab_others}99.0/23.2/32.4/\textbf{95.5} &\cellcolor{tab_ours}\textbf{99.4}/\textbf{43.0}/\textbf{47.9}/95.4\\

\midrule
cashew & 91.7/44.2/49.7/86.2 & 98.6/51.7/58.3/87.9 & \textbf{98.9}/\textbf{68.9}/\textbf{66.0}/84.1 &87.9/47.6/52.1/66.3 & 90.9/53.1/60.9/61.8 &94.3/46.8/51.4/87.8 &\cellcolor{tab_ours}97.1/64.5/62.4/\textbf{94.0}\\

\cellcolor{tab_others}chewinggum & \cellcolor{tab_others}98.7/59.9/61.7/76.9 & \cellcolor{tab_others}98.8/54.9/56.1/81.3 & \cellcolor{tab_others}97.9/26.8/29.8/78.3 &\cellcolor{tab_others}98.8/\textbf{86.9}/\textbf{81.0}/68.3 & \cellcolor{tab_others}94.7/11.9/25.8/59.5 &\cellcolor{tab_others}98.1/57.5/59.9/79.7 &\cellcolor{tab_ours}\textbf{99.1}/65.0/67.7/\textbf{88.1}\\

fryum & 97.0/47.6/51.5/93.4 & 95.9/34.0/40.6/76.2 &93.0/39.1/45.4/85.1 & 88.1/35.2/38.5/47.7 & \textbf{97.6}/\textbf{58.6}/\textbf{60.1}/81.3 &96.9/47.8/51.9/91.6 &\cellcolor{tab_ours}96.6/51.6/53.4/\textbf{93.5}\\

\cellcolor{tab_others}pipe\_fryum &\cellcolor{tab_others}99.1/56.8/58.8/\textbf{95.4} & \cellcolor{tab_others}98.9/50.2/57.7/91.5 & \cellcolor{tab_others}98.5/65.6/63.4/83.0 &\cellcolor{tab_others}98.9/78.8/72.7/45.9 & \cellcolor{tab_others}\textbf{99.4}/\textbf{72.7}/\textbf{69.9}/89.9 &\cellcolor{tab_others}99.1/53.5/58.5/95.1 &\cellcolor{tab_ours}99.2/64.3/65.1/95.2\\

\midrule
Mean & 98.1/38.0/42.6/91.8 & 95.9/21.0/27.0/75.6 & 96.8/34.7/37.8/81.4 &96.1/39.6/43.4/67.4 & 96.0/26.1/33.0/75.2 &98.5/39.4/44.0/91.0 &\cellcolor{tab_ours}\textbf{98.7}/\textbf{53.2}/\textbf{55.7}/\textbf{94.5}\\

\bottomrule
\end{tabular}
}
\label{tab:visapx}%
\end{table*}%

\begin{table*}[htbp]
\centering
\caption{Per-class performance on \textbf{Real-IAD} dataset for multi-class anomaly detection with AUROC/AP/$F_1$-max metrics.}
\resizebox{1.\linewidth}{!}{
\begin{tabular}{cccccccc}
\toprule
Method~$\rightarrow$ & RD4AD~\cite{deng2022anomaly} & UniAD~\cite{you2022unified} & SimpleNet~\cite{liu2023simplenet} & DeSTSeg~\cite{zhang2023destseg} & DiAD~\cite{he2024diffusion} & MambaAD & \cellcolor{tab_ours}Dinomaly \\
\cline{1-1}
Category~$\downarrow$ & CVPR'22 & NeurlPS'22 & CVPR'23 & CVPR'23 & AAAI'24 & Arxiv'24 & \cellcolor{tab_ours} Ours \\
\hline
audiojack & 76.2/63.2/60.8 & 81.4/76.6/64.9 & 58.4/44.2/50.9 & 81.1/72.6/64.5 & 76.5/54.3/65.7 & 84.2/76.5/67.4 & \cellcolor{tab_ours}\textbf{86.8}/\textbf{82.4}/\textbf{72.2} \\

\cellcolor{tab_others}bottle cap & \cellcolor{tab_others}89.5/86.3/81.0 & \cellcolor{tab_others}92.5/91.7/81.7 & \cellcolor{tab_others}54.1/47.6/60.3 & \cellcolor{tab_others}78.1/74.6/68.1 & \cellcolor{tab_others}91.6/\textbf{94.0}/\textbf{87.9} & \cellcolor{tab_others}\textbf{92.8}/92.0/82.1 & \cellcolor{tab_ours}89.9/86.7/81.2 \\

button battery & 73.3/78.9/76.1 & 75.9/81.6/76.3 & 52.5/60.5/72.4 & 86.7/89.2/83.5 & 80.5/71.3/70.6 & 79.8/85.3/77.8 & \cellcolor{tab_ours}86.6/88.9/82.1 \\

\cellcolor{tab_others}end cap & \cellcolor{tab_others}79.8/84.0/77.8 & \cellcolor{tab_others}80.9/86.1/78.0 & \cellcolor{tab_others}51.6/60.8/72.9 & \cellcolor{tab_others}77.9/81.1/77.1 & \cellcolor{tab_others}85.1/83.4/\textbf{84.8} & \cellcolor{tab_others}78.0/82.8/77.2 & \cellcolor{tab_ours}\textbf{87.0}/\textbf{87.5}/83.4 \\

eraser & 90.0/88.7/79.7 & \textbf{90.3}/\textbf{89.2}/\textbf{80.2} & 46.4/39.1/55.8 & 84.6/82.9/71.8 & 80.0/80.0/77.3 & 87.5/86.2/76.1 & \cellcolor{tab_ours}\textbf{90.3}/87.6/78.6 \\

\cellcolor{tab_others}fire hood & \cellcolor{tab_others}78.3/70.1/64.5 & \cellcolor{tab_others}80.6/74.8/66.4 & \cellcolor{tab_others}58.1/41.9/54.4 & \cellcolor{tab_others}81.7/72.4/67.7 & \cellcolor{tab_others}83.3/\textbf{81.7}/\textbf{80.5} & \cellcolor{tab_others}79.3/72.5/64.8 & \cellcolor{tab_ours}\textbf{83.8}/76.2/69.5 \\

mint & 65.8/63.1/64.8 & 67.0/66.6/64.6 & 52.4/50.3/63.7 & 58.4/55.8/63.7 & \textbf{76.7}/\textbf{76.7}/\textbf{76.0} & 70.1/70.8/65.5 & \cellcolor{tab_ours}73.1/72.0/67.7 \\

\cellcolor{tab_others}mounts & \cellcolor{tab_others}88.6/79.9/74.8 & \cellcolor{tab_others}87.6/77.3/77.2 & \cellcolor{tab_others}58.7/48.1/52.4 & \cellcolor{tab_others}74.7/56.5/63.1 & \cellcolor{tab_others}75.3/74.5/\textbf{82.5} & \cellcolor{tab_others}86.8/78.0/73.5 & \cellcolor{tab_ours}\textbf{90.4}/\textbf{84.2}/78.0 \\

pcb & 79.5/85.8/79.7 & 81.0/88.2/79.1 & 54.5/66.0/75.5 & 82.0/88.7/79.6 & 86.0/85.1/85.4 & 89.1/93.7/84.0 & \cellcolor{tab_ours}\textbf{92.0}/\textbf{95.3}/\textbf{87.0} \\

\cellcolor{tab_others}phone battery & \cellcolor{tab_others}87.5/83.3/77.1 & \cellcolor{tab_others}83.6/80.0/71.6 & \cellcolor{tab_others}51.6/43.8/58.0 & \cellcolor{tab_others}83.3/81.8/72.1 & \cellcolor{tab_others}82.3/77.7/75.9 & \cellcolor{tab_others}90.2/88.9/80.5 & \cellcolor{tab_ours}\textbf{92.9}/\textbf{91.6}/\textbf{82.5} \\

plastic nut & 80.3/68.0/64.4 & 80.0/69.2/63.7 & 59.2/40.3/51.8 & 83.1/75.4/66.5 & 71.9/58.2/65.6 & 87.1/80.7/70.7 & \cellcolor{tab_ours}\textbf{88.3}/\textbf{81.8}/\textbf{74.7} \\

\cellcolor{tab_others}plastic plug & \cellcolor{tab_others}81.9/74.3/68.8 & \cellcolor{tab_others}81.4/75.9/67.6 & \cellcolor{tab_others}48.2/38.4/54.6 & \cellcolor{tab_others}71.7/63.1/60.0 & \cellcolor{tab_others}88.7/\textbf{89.2}/\textbf{90.9} & \cellcolor{tab_others}85.7/82.2/72.6 & \cellcolor{tab_ours}\textbf{90.5}/86.4/78.6 \\

porcelain doll & 86.3/76.3/71.5 & 85.1/75.2/69.3 & 66.3/54.5/52.1 & 78.7/66.2/64.3 & 72.6/66.8/65.2 & \textbf{88.0}/\textbf{82.2}/\textbf{74.1} & \cellcolor{tab_ours}85.1/73.3/69.6 \\

\cellcolor{tab_others}regulator & \cellcolor{tab_others}66.9/48.8/47.7 & \cellcolor{tab_others}56.9/41.5/44.5 & \cellcolor{tab_others}50.5/29.0/43.9 & \cellcolor{tab_others}79.2/63.5/56.9 & \cellcolor{tab_others}72.1/71.4/\textbf{78.2} & \cellcolor{tab_others}69.7/58.7/50.4 & \cellcolor{tab_ours}\textbf{85.2}/\textbf{78.9}/69.8 \\

rolled strip base & 97.5/98.7/94.7 & 98.7/99.3/96.5 & 59.0/75.7/79.8 & 96.5/98.2/93.0 & 68.4/55.9/56.8 & 98.0/99.0/95.0 & \cellcolor{tab_ours}\textbf{99.2}/\textbf{99.6}/\textbf{97.1} \\

\cellcolor{tab_others}sim card set & \cellcolor{tab_others}91.6/91.8/84.8 & \cellcolor{tab_others}89.7/90.3/83.2 & \cellcolor{tab_others}63.1/69.7/70.8 & \cellcolor{tab_others}95.5/96.2/\textbf{89.2} & \cellcolor{tab_others}72.6/53.7/61.5 & \cellcolor{tab_others}94.4/95.1/87.2 & \cellcolor{tab_ours}\textbf{95.8}/\textbf{96.3}/88.8 \\

switch & 84.3/87.2/77.9 & 85.5/88.6/78.4 & 62.2/66.8/68.6 & 90.1/92.8/83.1 & 73.4/49.4/61.2 & 91.7/94.0/85.4 & \cellcolor{tab_ours}\textbf{97.8}/\textbf{98.1}/\textbf{93.3} \\

\cellcolor{tab_others}tape & \cellcolor{tab_others}96.0/95.1/87.6 & \cellcolor{tab_others}\textbf{97.2}/\textbf{96.2}/\textbf{89.4} & \cellcolor{tab_others}49.9/41.1/54.5 & \cellcolor{tab_others}94.5/93.4/85.9 & \cellcolor{tab_others}73.9/57.8/66.1 & \cellcolor{tab_others}96.8/95.9/89.3 & \cellcolor{tab_ours}96.9/95.0/88.8 \\

terminalblock & 89.4/89.7/83.1 & 87.5/89.1/81.0 & 59.8/64.7/68.8 & 83.1/86.2/76.6 & 62.1/36.4/47.8 & 96.1/96.8/90.0 & \cellcolor{tab_ours}\textbf{96.7}/\textbf{97.4}/\textbf{91.1} \\

\cellcolor{tab_others}toothbrush & \cellcolor{tab_others}82.0/83.8/77.2 & \cellcolor{tab_others}78.4/80.1/75.6 & \cellcolor{tab_others}65.9/70.0/70.1 & \cellcolor{tab_others}83.7/85.3/79.0 & \cellcolor{tab_others}\textbf{91.2}/\textbf{93.7}/\textbf{90.9} & \cellcolor{tab_others}85.1/86.2/80.3 & \cellcolor{tab_ours}90.4/91.9/83.4\\

toy & 69.4/74.2/75.9 & 68.4/75.1/74.8 & 57.8/64.4/73.4 & 70.3/74.8/75.4 & 66.2/57.3/59.8 & 83.0/87.5/79.6 & \cellcolor{tab_ours}85.6/89.1/81.9 \\

\cellcolor{tab_others}toy brick & \cellcolor{tab_others}63.6/56.1/59.0 & \cellcolor{tab_others}\textbf{77.0}/\textbf{71.1}/66.2 & \cellcolor{tab_others}58.3/49.7/58.2 & \cellcolor{tab_others}73.2/68.7/\textbf{63.3} & \cellcolor{tab_others}68.4/45.3/55.9 & \cellcolor{tab_others}70.5/63.7/61.6 & \cellcolor{tab_ours}72.3/65.1/63.4 \\

transistor1 & 91.0/94.0/85.1 & 93.7/95.9/88.9 & 62.2/69.2/72.1 & 90.2/92.1/84.6 & 73.1/63.1/62.7 & 94.4/96.0/89.0 & \cellcolor{tab_ours}\textbf{97.4}/\textbf{98.2}/\textbf{93.1} \\

\cellcolor{tab_others}u block & \cellcolor{tab_others}89.5/85.0/74.2 & \cellcolor{tab_others}88.8/84.2/75.5 & \cellcolor{tab_others}62.4/48.4/51.8 & \cellcolor{tab_others}80.1/73.9/64.3 & \cellcolor{tab_others}75.2/68.4/67.9 & \cellcolor{tab_others}89.7/\textbf{85.7}/\textbf{75.3} & \cellcolor{tab_ours}\textbf{89.9}/84.0/75.2 \\

usb & 84.9/84.3/75.1 & 78.7/79.4/69.1 & 57.0/55.3/62.9 & 87.8/88.0/78.3 & 58.9/37.4/45.7 & \textbf{92.0}/\textbf{92.2}/\textbf{84.5} & \cellcolor{tab_ours}\textbf{92.0}/91.6/83.3 \\

\cellcolor{tab_others}usb adaptor & \cellcolor{tab_others}71.1/61.4/62.2 & \cellcolor{tab_others}76.8/71.3/64.9 & \cellcolor{tab_others}47.5/38.4/56.5 & \cellcolor{tab_others}80.1/\textbf{74.9}/67.4 & \cellcolor{tab_others}76.9/60.2/67.2 & \cellcolor{tab_others}79.4/76.0/66.3 & \cellcolor{tab_ours}\textbf{81.5}/74.5/\textbf{69.4} \\

vcpill & 85.1/80.3/72.4 & 87.1/84.0/74.7 & 59.0/48.7/56.4 & 83.8/81.5/69.9 & 64.1/40.4/56.2 & 88.3/87.7/77.4 & \cellcolor{tab_ours}\textbf{92.0}/\textbf{91.2}/\textbf{82.0} \\

\cellcolor{tab_others}wooden beads & \cellcolor{tab_others}81.2/78.9/70.9 & \cellcolor{tab_others}78.4/77.2/67.8 & \cellcolor{tab_others}55.1/52.0/60.2 & \cellcolor{tab_others}82.4/78.5/73.0 & \cellcolor{tab_others}62.1/56.4/65.9 & \cellcolor{tab_others}82.5/81.7/71.8 & \cellcolor{tab_ours}\textbf{87.3}/\textbf{85.8}/\textbf{77.4} \\

woodstick & 76.9/61.2/58.1 & 80.8/72.6/63.6 & 58.2/35.6/45.2 & 80.4/69.2/60.3 & 74.1/66.0/62.1 & 80.4/69.0/63.4 & \cellcolor{tab_ours}\textbf{84.0}/\textbf{73.3}/\textbf{65.6} \\

\cellcolor{tab_others}zipper & \cellcolor{tab_others}95.3/97.2/91.2 & \cellcolor{tab_others}98.2/98.9/95.3 & \cellcolor{tab_others}77.2/86.7/77.6 & \cellcolor{tab_others}96.9/98.1/93.5 & \cellcolor{tab_others}86.0/87.0/84.0 & \cellcolor{tab_others}\textbf{99.2}/\textbf{99.6}/\textbf{96.9} & \cellcolor{tab_ours}99.1/99.5/96.5 \\

\midrule
Mean & 82.4/79.0/73.9 & 83.0/80.9/74.3 & 57.2/53.4/61.5 & 82.3/79.2/73.2 & 75.6/66.4/69.9 & 86.3/84.6/77.0 & \cellcolor{tab_ours}\textbf{89.3}/\textbf{86.8}/\textbf{80.2} \\
\bottomrule
\end{tabular}
}
\label{tab:realiadsp}
\end{table*}

\begin{table*}[htp!]
\centering
\caption{Per-class performance on \textbf{Real-IAD} dataset for multi-class anomaly localization with AUROC/AP/$F_1$-max/AUPRO metrics.}
\resizebox{1.\linewidth}{!}{
\begin{tabular}{cccccccc}
\toprule
Method~$\rightarrow$ & RD4AD~\cite{deng2022anomaly} & UniAD~\cite{you2022unified} & SimpleNet~\cite{liu2023simplenet} & DeSTSeg~\cite{zhang2023destseg} & DiAD~\cite{he2024diffusion} & MambaAD~\cite{he2024mambaad} & \cellcolor{tab_ours}Dinomaly \\
\cline{1-1}
Category~$\downarrow$ & CVPR'22 & NeurlPS'22 & CVPR'23 & CVPR'23 & AAAI'24 & Arxiv'24 & \cellcolor{tab_ours} Ours \\
\hline
audiojack & 96.6/12.8/22.1/79.6 & 97.6/20.0/31.0/83.7 & 74.4/ 0.9/ 4.8/38.0 & 95.5/25.4/31.9/52.6 & 91.6/ 1.0/ 3.9/63.3 & 97.7/21.6/29.5/83.9 & \cellcolor{tab_ours}\textbf{98.7}/\textbf{48.1}/\textbf{54.5}/\textbf{91.7} \\

\cellcolor{tab_others}bottle cap & \cellcolor{tab_others}99.5/18.9/29.9/95.7 & \cellcolor{tab_others}99.5/19.4/29.6/96.0 & \cellcolor{tab_others}85.3/ 2.3/ 5.7/45.1 & \cellcolor{tab_others}94.5/25.3/31.1/25.3 & \cellcolor{tab_others}94.6/ 4.9/11.4/73.0 & \textbf{99.7}/30.6/34.6/97.2 & \cellcolor{tab_ours}\textbf{99.7}/\textbf{32.4}/\textbf{36.7}/\textbf{98.1} \\

button battery& 97.6/33.8/37.8/86.5 & 96.7/28.5/34.4/77.5 & 75.9/ 3.2/ 6.6/40.5 & 98.3/\textbf{63.9}/\textbf{60.4}/36.9 & 84.1/ 1.4/ 5.3/66.9 & 98.1/46.7/49.5/86.2 & \cellcolor{tab_ours}\textbf{99.1}/46.9/56.7/\textbf{92.9} \\

\cellcolor{tab_others}end cap & \cellcolor{tab_others}96.7/12.5/22.5/89.2 & \cellcolor{tab_others}95.8/ 8.8/17.4/85.4 & \cellcolor{tab_others}63.1/ 0.5/ 2.8/25.7 & \cellcolor{tab_others}89.6/14.4/22.7/29.5 & \cellcolor{tab_others}81.3/ 2.0/ 6.9/38.2 & 97.0/12.0/19.6/89.4 & \cellcolor{tab_ours}\textbf{99.1}/\textbf{26.2}/\textbf{32.9}/\textbf{96.0} \\

eraser & \textbf{99.5}/30.8/36.7/96.0 & 99.3/24.4/30.9/94.1 & 80.6/ 2.7/ 7.1/42.8 & 95.8/52.7/53.9/46.7 & 91.1/ 7.7/15.4/67.5 & 99.2/30.2/38.3/93.7 & \cellcolor{tab_ours}\textbf{99.5}/\textbf{39.6}/\textbf{43.3}/\textbf{96.4} \\

\cellcolor{tab_others}fire hood & \cellcolor{tab_others}98.9/27.7/35.2/87.9 & \cellcolor{tab_others}98.6/23.4/32.2/85.3 & \cellcolor{tab_others}70.5/ 0.3/ 2.2/25.3 & \cellcolor{tab_others}97.3/27.1/35.3/34.7 & \cellcolor{tab_others}91.8/ 3.2/ 9.2/66.7 & 98.7/25.1/31.3/86.3 & \cellcolor{tab_ours}\textbf{99.3}/\textbf{38.4}/\textbf{42.7}/\textbf{93.0} \\

mint & 95.0/11.7/23.0/72.3 & 94.4/ 7.7/18.1/62.3 & 79.9/ 0.9/ 3.6/43.3 & 84.1/10.3/22.4/ 9.9 & 91.1/ 5.7/11.6/64.2 & 96.5/15.9/27.0/72.6 & \cellcolor{tab_ours}\textbf{96.9}/\textbf{22.0}/\textbf{32.5}/\textbf{77.6} \\

\cellcolor{tab_others}mounts & \cellcolor{tab_others}99.3/30.6/37.1/94.9 & \cellcolor{tab_others}\textbf{99.4}/28.0/32.8/95.2 & \cellcolor{tab_others}80.5/ 2.2/ 6.8/46.1 & \cellcolor{tab_others}94.2/30.0/41.3/43.3 & \cellcolor{tab_others}84.3/ 0.4/ 1.1/48.8 & 99.2/31.4/35.4/93.5 & \cellcolor{tab_ours}\textbf{99.4}/\textbf{39.9}/\textbf{44.3}/\textbf{95.6} \\

pcb & 97.5/15.8/24.3/88.3 & 97.0/18.5/28.1/81.6 & 78.0/ 1.4/ 4.3/41.3 & 97.2/37.1/40.4/48.8 & 92.0/ 3.7/ 7.4/66.5 & 99.2/46.3/50.4/93.1 & \cellcolor{tab_ours}\textbf{99.3}/\textbf{55.0}/\textbf{56.3}/\textbf{95.7} \\

\cellcolor{tab_others}phone battery & \cellcolor{tab_others}77.3/22.6/31.7/94.5 & \cellcolor{tab_others}85.5/11.2/21.6/88.5 & \cellcolor{tab_others}43.4/ 0.1/ 0.9/11.8 & \cellcolor{tab_others}79.5/25.6/33.8/39.5 & \cellcolor{tab_others}96.8/ 5.3/11.4/85.4 & 99.4/36.3/41.3/95.3 & \cellcolor{tab_ours}\textbf{99.7}/\textbf{51.6}/\textbf{54.2}/\textbf{96.8}\\

\cellcolor{tab_others}phone battery& \cellcolor{tab_others}77.3/22.6/31.7/94.5 &\cellcolor{tab_others}85.5/11.2/21.6/88.5 &\cellcolor{tab_others}43.4/ 0.1/ 0.9/11.8 &\cellcolor{tab_others}79.5/25.6/33.8/39.5 &\cellcolor{tab_others}96.8/5.3/11.4/85.4 &99.4/36.3/41.3/95.3 & \cellcolor{tab_ours}\textbf{99.7}/\textbf{51.6}/\textbf{54.2}/\textbf{96.8}\\

plastic nut& 98.8/21.1/29.6/91.0 &98.4/20.6/27.1/88.9 &77.4/ 0.6/ 3.6/41.5 &96.5/44.8/45.7/38.4 &81.1/ 0.4/ 3.4/38.6 &99.4/33.1/37.3/96.1 & \cellcolor{tab_ours}\textbf{99.7}/\textbf{41.0}/\textbf{45.0}/\textbf{97.4}\\

\cellcolor{tab_others}plastic plug& \cellcolor{tab_others}99.1/20.5/28.4/94.9 &\cellcolor{tab_others}98.6/17.4/26.1/90.3 &\cellcolor{tab_others}78.6/ 0.7/ 1.9/38.8 &\cellcolor{tab_others}91.9/20.1/27.3/21.0 &\cellcolor{tab_others}92.9/ 8.7/15.0/66.1 &99.0/24.2/31.7/91.5 & \cellcolor{tab_ours}\textbf{99.4}/\textbf{31.7}/\textbf{37.2}/\textbf{96.4}\\

porcelain doll& 99.2/24.8/34.6/95.7 &98.7/14.1/24.5/93.2 &81.8/ 2.0/ 6.4/47.0 &93.1/35.9/40.3/24.8 &93.1/ 1.4/ 4.8/70.4 &99.2/\textbf{31.3}/\textbf{36.6}/95.4 & \cellcolor{tab_ours}\textbf{99.3}/27.9/33.9/\textbf{96.0}\\

\cellcolor{tab_others}regulator& \cellcolor{tab_others}98.0/7.8/16.1/88.6 &\cellcolor{tab_others}95.5/9.1/17.4/76.1 &\cellcolor{tab_others}76.6/0.1/0.6/38.1 &\cellcolor{tab_others}88.8/18.9/23.6/17.5 &\cellcolor{tab_others}84.2/0.4/1.5/44.4 &97.6/20.6/29.8/87.0 & \cellcolor{tab_ours}\textbf{99.3}/\textbf{42.2}/\textbf{48.9}/\textbf{95.6}\\

rolled strip base& \textbf{99.7}/31.4/39.9/98.4 &99.6/20.7/32.2/97.8 &80.5/ 1.7/ 5.1/52.1 &99.2/\textbf{48.7}/\textbf{50.1}/55.5 &87.7/ 0.6/ 3.2/63.4 &99.7/37.4/42.5/98.8 & \cellcolor{tab_ours}\textbf{99.7}/41.6/45.5/\textbf{98.5}\\

\cellcolor{tab_others}sim card set& \cellcolor{tab_others}98.5/40.2/44.2/89.5 &\cellcolor{tab_others}97.9/31.6/39.8/85.0 &\cellcolor{tab_others}71.0/ 6.8/14.3/30.8 &\cellcolor{tab_others}\textbf{99.1}/\textbf{65.5}/\textbf{62.1}/73.9 &\cellcolor{tab_others}89.9/ 1.7/ 5.8/60.4 &98.8/51.1/50.6/89.4 & \cellcolor{tab_ours}99.0/52.1/52.9/\textbf{90.9}\\

switch& 94.4/18.9/26.6/90.9 &98.1/33.8/40.6/90.7 &71.7/ 3.7/ 9.3/44.2 &97.4/57.6/55.6/44.7 &90.5/ 1.4/ 5.3/64.2 &\textbf{98.2}/39.9/45.4/92.9 & \cellcolor{tab_ours}96.7/\textbf{62.3}/\textbf{63.6}/\textbf{95.9}\\

\cellcolor{tab_others}tape& \cellcolor{tab_others}99.7/42.4/47.8/98.4 &\cellcolor{tab_others}99.7/29.2/36.9/97.5 &\cellcolor{tab_others}77.5/ 1.2/ 3.9/41.4 &\cellcolor{tab_others}99.0/61.7/57.6/48.2 &\cellcolor{tab_others}81.7/ 0.4/ 2.7/47.3 &\textbf{99.8}/47.1/48.2/98.0 & \cellcolor{tab_ours}\textbf{99.8}/\textbf{54.0}/\textbf{55.8}/98.8\\

terminalblock& 99.5/27.4/35.8/97.6 &99.2/23.1/30.5/94.4 &87.0/ 0.8/ 3.6/54.8 &96.6/40.6/44.1/34.8 &75.5/ 0.1/ 1.1/38.5 &\textbf{99.8}/35.3/39.7/98.2 & \cellcolor{tab_ours}\textbf{99.8}/\textbf{48.0}/\textbf{50.7}/\textbf{98.8}\\

\cellcolor{tab_others}toothbrush& \cellcolor{tab_others}96.9/26.1/34.2/88.7 &\cellcolor{tab_others}95.7/16.4/25.3/84.3 &\cellcolor{tab_others}84.7/ 7.2/14.8/52.6 &\cellcolor{tab_others}94.3/30.0/37.3/42.8&\cellcolor{tab_others}82.0/ 1.9/ 6.6/54.5&\textbf{97.5}/27.8/36.7/\textbf{91.4} & \cellcolor{tab_ours}96.9/\textbf{38.3}/\textbf{43.9}/90.4\\

toy & 95.2/ 5.1/12.8/82.3 & 93.4/ 4.6/12.4/70.5 &67.7/ 0.1/ 0.4/25.0 &86.3/ 8.1/15.9/16.4 &82.1/ 1.1/ 4.2/50.3 &\textbf{96.0}/16.4/25.8/86.3 & \cellcolor{tab_ours}94.9/\textbf{22.5}/\textbf{32.1}/\textbf{91.0}\\

\cellcolor{tab_others}toy brick& \cellcolor{tab_others}96.4/16.0/24.6/75.3 &\cellcolor{tab_others}\textbf{97.4}/17.1/27.6/\textbf{81.3} &\cellcolor{tab_others}86.5/ 5.2/11.1/56.3 &\cellcolor{tab_others}94.7/24.6/30.8/45.5 &\cellcolor{tab_others}93.5/ 3.1/ 8.1/66.4 &96.6/18.0/25.8/74.7 & \cellcolor{tab_ours}96.8/\textbf{27.9}/\textbf{34.0}/76.6\\

transistor1& 99.1/29.6/35.5/95.1 &98.9/25.6/33.2/94.3 &71.7/ 5.1/11.3/35.3 &97.3/43.8/44.5/45.4 &88.6/ 7.2/15.3/58.1 &99.4/39.4/40.0/96.5 & \cellcolor{tab_ours}\textbf{99.6}/\textbf{53.5}/\textbf{53.3}/\textbf{97.8}\\

\cellcolor{tab_others}u block& \cellcolor{tab_others}99.6/40.5/45.2/96.9 &\cellcolor{tab_others}99.3/22.3/29.6/94.3 &\cellcolor{tab_others}76.2/ 4.8/12.2/34.0 &\cellcolor{tab_others}96.9/\textbf{57.1}/\textbf{55.7}/38.5 &\cellcolor{tab_others}88.8/ 1.6/ 5.4/54.2 &\textbf{99.5}/37.8/46.1/95.4 & \cellcolor{tab_ours}\textbf{99.5}/41.8/45.6/\textbf{96.8}\\

usb& 98.1/26.4/35.2/91.0 &97.9/20.6/31.7/85.3 &81.1/ 1.5/ 4.9/52.4 &98.4/42.2/47.7/57.1 &78.0/ 1.0/ 3.1/28.0 &\textbf{99.2}/39.1/44.4/95.2 & \cellcolor{tab_ours}\textbf{99.2}/\textbf{45.0}/\textbf{48.7}/\textbf{97.5}\\

\cellcolor{tab_others}usb adaptor& \cellcolor{tab_others}94.5/ 9.8/17.9/73.1 &\cellcolor{tab_others}96.6/10.5/19.0/78.4 &\cellcolor{tab_others}67.9/ 0.2/ 1.3/28.9 &\cellcolor{tab_others}94.9/\textbf{25.5}/\textbf{34.9}/36.4 &\cellcolor{tab_others}94.0/ 2.3/ 6.6/75.5 &97.3/15.3/22.6/82.5 & \cellcolor{tab_ours}\textbf{98.7}/23.7/32.7/\textbf{91.0}\\

vcpill& 98.3/43.1/48.6/88.7 &99.1/40.7/43.0/91.3 &68.2/ 1.1/ 3.3/22.0 &97.1/64.7/62.3/42.3 &90.2/ 1.3/ 5.2/60.8 &98.7/50.2/54.5/89.3 & \cellcolor{tab_ours}\textbf{99.1}/\textbf{66.4}/\textbf{66.7}/\textbf{93.7}\\

\cellcolor{tab_others}wooden beads&	\cellcolor{tab_others}98.0/27.1/34.7/85.7 &\cellcolor{tab_others}97.6/16.5/23.6/84.6 &\cellcolor{tab_others}68.1/ \;2.4/ \;6.0/28.3 &\cellcolor{tab_others}94.7/38.9/42.9/39.4 &\cellcolor{tab_others}85.0/ \;1.1/ \;4.7/45.6 &\cellcolor{tab_others}98.0/32.6/39.8/84.5 & \cellcolor{tab_ours}\textbf{99.1}/\textbf{45.8}/\textbf{50.1}/\textbf{90.5}\\

woodstick& 97.8/30.7/38.4/85.0 &94.0/36.2/44.3/77.2 &76.1/ 1.4/ 6.0/32.0 &97.9/\textbf{60.3}/\textbf{60.0}/51.0 &90.9/ 2.6/ 8.0/60.7 &97.7/40.1/44.9/82.7 & \cellcolor{tab_ours}\textbf{99.0}/50.9/52.1/\textbf{90.4}\\

\cellcolor{tab_others}zipper& \cellcolor{tab_others}99.1/44.7/50.2/96.3 &\cellcolor{tab_others}98.4/32.5/36.1/95.1 &\cellcolor{tab_others}89.9/23.3/31.2/55.5 &\cellcolor{tab_others}98.2/35.3/39.0/78.5 &\cellcolor{tab_others}90.2/12.5/18.8/53.5 &99.3/58.2/61.3/97.6 & \cellcolor{tab_ours}\textbf{99.3}/\textbf{67.2}/\textbf{66.5}/\textbf{97.8}\\
\hline
Mean& 97.3/25.0/32.7/89.6 &97.3/21.1/29.2/86.7 &75.7/ 2.8/ 6.5/39.0 &94.6/37.9/41.7/40.6 &88.0/ 2.9/ 7.1/58.1 &98.5/33.0/38.7/90.5 & \cellcolor{tab_ours}\textbf{98.8}/\textbf{42.8}/\textbf{47.1}/\textbf{93.9}\\

\bottomrule
\end{tabular}
}
\label{tab:realiadpx}
\end{table*}

